\pdfoutput=1

\documentclass[10pt,twocolumn,letterpaper]{article}

    \usepackage[pagenumbers]{cvpr} %

\definecolor{cvprblue}{rgb}{0.21,0.49,0.74}
\usepackage[pagebackref,breaklinks,colorlinks,allcolors=cvprblue]{hyperref}
\usepackage{graphicx}
\usepackage{booktabs}
\usepackage{colortbl}
\usepackage{multirow}
\usepackage{float}
\usepackage{tikz}
\usetikzlibrary{calc, shapes, arrows.meta, positioning, fit, backgrounds, decorations.pathreplacing}
\usepackage{amsmath, amssymb}
\usepackage{caption}
\usepackage{svg}

\newcommand{\supref}[1]{\cref{#1}}
\newcommand{\mainref}[1]{\cref{#1}}
\makeatletter
\let\@oldmaketitle\@maketitle
\renewcommand{\@maketitle}{\@oldmaketitle
  \begin{center}
    \vspace{-25pt}\centering\resizebox{\textwidth}{!}{%
  $\vcenter{\hbox{\includegraphics[height=3.3cm]{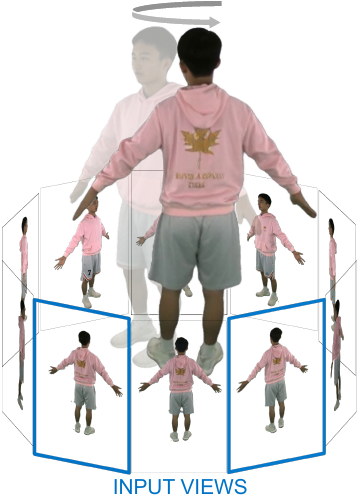}}}$ \hspace{0.2cm}%
  $\vcenter{\hbox{\includegraphics[height=3.3cm]{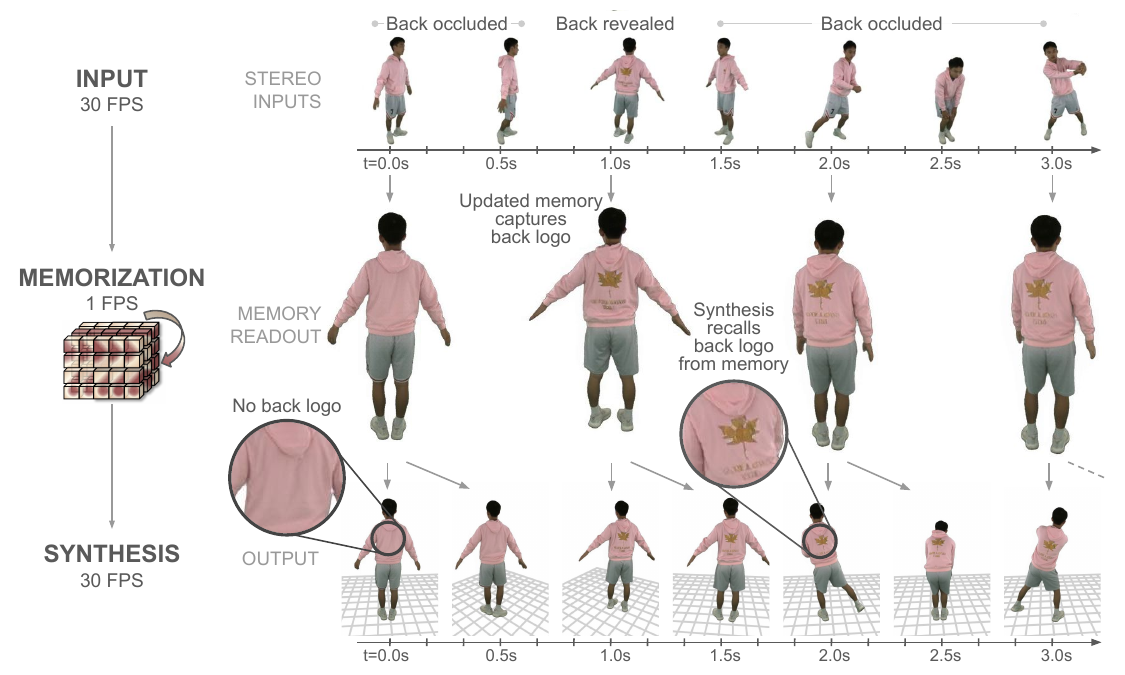}}}$
  \hspace{0.2cm}%
  $\vcenter{\hbox{\includegraphics[height=3.3cm]{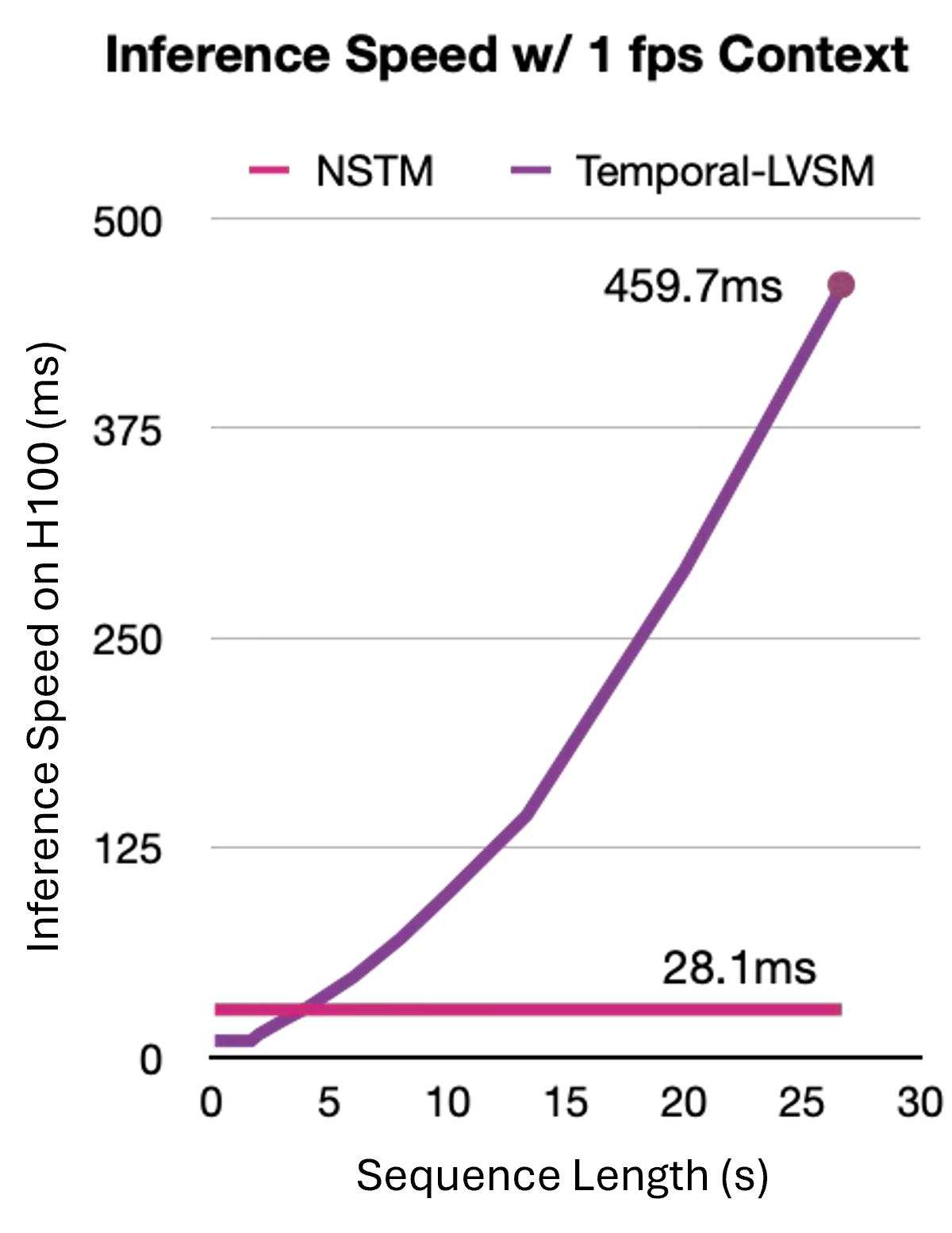}}}$%
}

\captionof{figure}{\textbf{Neural Space-Time Memory (NSTM)} processes 30 FPS multi-view video streams, updating its memory at 1 FPS while synthesizing novel views in real time. 
\textbf{Middle:} NSTM captures back logo as subject rotates (see memory readout), faithfully recalling it later from memory given only frontal views. \textbf{Right:} While full self-attention in typical feed-forward models (Temporal-LVSM see~\cref{subsec:efficiency}) scales quadratically with history sequence length, NSTM scales at $\mathcal{O}(1)$ complexity, achieving $256\times256$ amortized real-time inference.
}

\vspace{-10pt}
\label{fig:teaser}

  \end{center}
  \vspace{1em} %
}
\makeatother
\makeatletter
\def\NAT@cmprs{\z@}   %
\makeatother

\title{Online Neural Space Time Memory for Dynamic Novel View Synthesis}

\author{
Baback Elmieh$^{1,2}$\thanks{This work was done while Baback was a student researcher at Google.} \quad
Lynn Tsai$^1$ \quad
Zeman Li$^1$ \quad
Srinivas Kaza$^1$ \quad
Tiancheng Sun$^1$ \\
Gabor Csapo$^1$ \quad
Ali Behrouz$^1$ \quad
Yuan Deng$^1$ \quad
Stephen Lombardi$^1$ \\
Steven M. Seitz$^1$ \quad
Xuan Luo$^1$ \\
\\
$^1$Google \quad $^2$University of Washington
}

\begin{document}
\maketitle

\definecolor{best}{HTML}{FFC7CE}
\definecolor{second}{HTML}{FFEB84}
\definecolor{third}{HTML}{FFFFC7}

\begin{abstract}

\noindent Online novel view synthesis from multi-view streaming videos faces a fundamental trade-off: maintaining a persistent, long-horizon memory to reconstruct temporarily occluded regions while operating under strict real-time constraints. While Test-Time Training (TTT) offers a powerful memory mechanism, standard models mandate gradient-based memory updates at every frame to adapt to the changing motion in dynamic scenes. The computational cost of heavy memory updates precludes real-time application and can lead to instability over long contexts. Given that memory updates are more demanding than memory application and video content is largely redundant, we propose to decouple the frequencies of these two processes. Our approach performs periodic memory updates while applying the memory on a per-frame basis, using cross-view attention to manage deformations between the prior memory state and the current frame. To lock in the historical context, we introduce two critical mechanisms: an auxiliary \emph{Memory Loss} that forces persistent internalization of the scene, and a \emph{Memory Caching} strategy that regularizes active weights against catastrophic drift. Our method demonstrates state-of-the-art minute-scale memory persistence in online dynamic human scenes at amortized real-time speed.

    \noindent Please see our project page for extensive demos and results: \href{https://nst-mem.github.io}{nst-mem.github.io}.
    \vspace{-15px}

\end{abstract}

\section{Introduction}
\label{sec:intro}

Online novel view synthesis (NVS) of dynamic scenes from multi-view streaming videos is a highly sought-after goal in computer vision, underpinning many telepresence applications such as 3D teleconferencing, holographic communication, and live free-viewpoint broadcasts. However, sparse camera setups and occlusions often make observations incomplete 
at a single instant. For example, synthesizing a person's back when current inputs provide only frontal views is challenging without historical context (\cref{fig:teaser}). 
Solving this problem requires a \emph{space-time memory} -- akin to how humans integrate information across space and time, naturally recalling previously observed, now-occluded visual details.
However, current online dynamic NVS methods struggle to maintain a persistent memory over a long time without sacrificing real-time speed.

\noindent In this paper, we present a system for online dynamic NVS from multi-view videos that can retain minute-long historical context to perform grounded synthesis of momentarily occluded regions while operating under amortized real-time constraints. At the core of our approach is a \emph{space-time memory} mechanism based on \emph{Test-Time Training (TTT)}. We leverage TTT for its expressive memory capacity and, crucially, its linear scalability -- bypassing the prohibitive quadratic computational cost of full self-attention heavily used in typical feed-forward NVS methods (\cref{fig:teaser}). Nevertheless, while TTT has shown promise in static 3D environments, adapting it to continuous, streaming dynamic scenes in real time poses several challenges, as follows.

\begin{figure*}[t]
  \vspace{-20pt}
  \centering
  \includegraphics[width=1.0\textwidth]{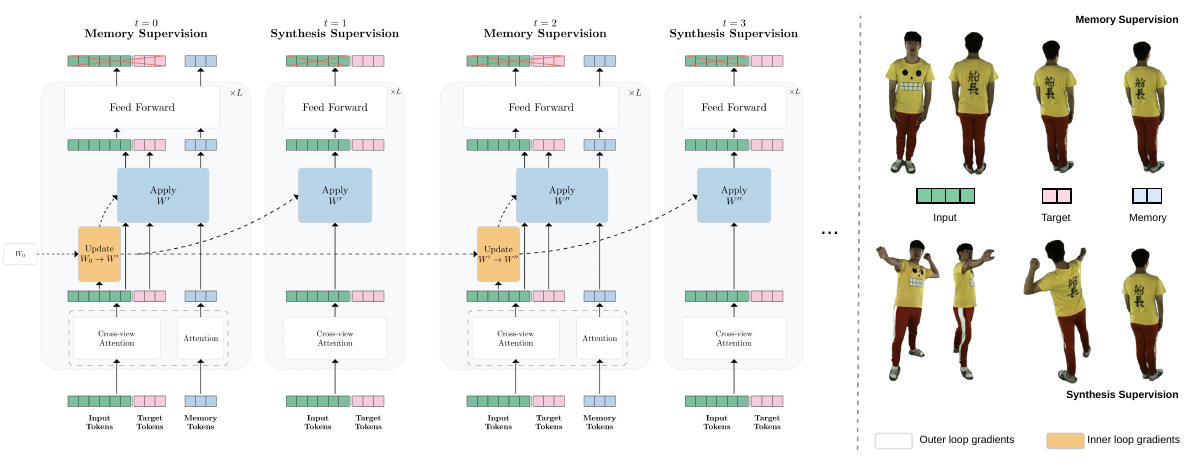}
  \vspace{-20pt}
  \caption{\textbf{NSTM Training Scheme.}  
  \textbf{Left:} Our alternating training scheme explicitly enforces memorization. In the \emph{Memory Supervision} step, isolated \emph{memory tokens} (target camera rays) perform strict self-attention --- sharing weights with the \emph{cross-view attention} layer --- without attending to input views. Once the memory is updated by the inputs, these isolated tokens query the memory to reconstruct the target view. This forces a pure memory readout, ensuring context internalization. During \emph{Synthesis Supervision}, we perform self-attention across input and target tokens to capture ongoing motion, and then query the previously updated memory state. \textbf{Right:} Visualizations of the inputs, targets, and memory in both steps. Crucially, the memory state predates the current observation during synthesis: here the memory holds a standing pose (arms down, bottom-right) while the current inputs show a throwing pose (arms raised, bottom-left). Our \emph{cross-view attention} resolves this misalignment, transferring the memorized shirt-back pattern onto the current pose.
  }
  \label{fig:training_regime}
  \vspace{-15pt}
\end{figure*}

\noindent\textbf{Speed.} Real-time dynamic NVS via TTT faces an architectural bottleneck. Standard TTT gains efficiency through \emph{large-chunk} processing and per-image attention \cite{zhang2025test_lact, zhang2026loger}. However, this per-image attention forces the model to rely entirely on the memory module to synthesize target views. Thus tracking continuous motion requires per-timestep (\emph{small-chunk}) memory updates, breaking the large-chunk efficiency. This full per-timestep TTT step (memory update and apply) takes \textbf{58ms} at $256\times256$ on an H100, precluding real-time deployment. Conversely, pure memory apply requires only \textbf{27ms} (37 FPS). Exploiting this computational asymmetry and video redundancy, we propose to \emph{decouple} their frequencies (\cref{fig:teaser}): updating memory (\emph{memorization}) periodically while applying it (\emph{synthesis}) per frame. To resolve the resulting motion misalignment between the periodically updated memory and the current frame, we utilize \emph{cross-view attention} (\cref{fig:training_regime}) -- it allows target views to directly attend to current inputs, fusing ongoing motion with past memory context to achieve amortized real-time speed.

\noindent\textbf{Memory Persistence and Supervision Balance.} Decoupling synthesis and memorization frequencies requires careful design of the training objective. This is because occlusion signals in natural videos are spatially and temporally sparse. Jointly training the separate memorization and synthesis tasks on this sparse data distribution results in an underconstrained problem, causing the network to over-rely on processing current, incomplete observation, effectively bypassing the memory module and forgetting the past (\cref{sec:Ablation}). To resolve this, we disentangle the training objectives and propose an auxiliary \textbf{\emph{memory loss}}. During the \emph{Memorization Step}, we query the memory using only the target view camera rays (memory tokens in \cref{fig:training_regime}), deliberately preventing any attention over the current input views. By isolating the readout, we force the memory module to reconstruct the scene entirely from its internalized historical context. 
By explicitly supervising the memory's independent reconstruction capabilities, we force the memory module to persistently internalize the scene context.

\noindent\textbf{Long-Term Stability.} Continuous memory updates in TTT can lead to instability over long time horizons. Our periodic update scheme natively mitigates this by reducing the overall update frequency. To further enhance stability, we adopt a \textbf{\emph{memory caching}} strategy~\cite{behrouzmemorycaching2026} that aggregates historical memory checkpoints, stabilizing the parameter space to handle minute-long streams. Crucially, modeling deformation in the Synthesis Step enables memory caching to handle changing motion (\cref{sec:Ablation}). Finally, we identify that the standard dot-product inner loss used in state-of-the-art TTT works \cite{zhang2025test_lact, zhang2026loger, jin2026zipmap} causes magnitude inflation over many update steps and show that we can stabilize the framework by replacing it with an $L_2$ objective~\cite{behrouz2025atlaslearningoptimallymemorize, behrouz2026_ALLCONNECTED}.

\noindent While our method is not human-specific, we evaluate in the human telepresence setting with the large-scale MVHumanNet++ dataset~\cite{li2025mvhumannet++}, as human motion represents one of the most prominent and challenging applications of online dynamic NVS from multi-view videos. Despite the dataset's diverse and complex motion and garments, NSTM successfully delivers high-fidelity synthesis and stable, minute-long persistent memorization at amortized real-time speeds.

\noindent Our key contributions are summarized as follows:

\begin{itemize}
    \item \textbf{Real-Time, Long-Context Dynamic NVS:} We propose Neural Space-Time Memory (NSTM), the first online framework for dynamic novel view synthesis from multi-view videos that sustains minute-long persistent memory while operating in amortized real-time.
    
    \item \textbf{Decoupled Memorization and Deformation:} We overcome the TTT computational bottleneck by structurally decoupling the frequencies between periodic, heavy memory updates and continuous, transformer-driven per-frame synthesis.
    
    \item \textbf{Disentangled Memory Supervision:} We introduce an alternating \emph{memorization} and \emph{synthesis} training regime and an auxiliary \textbf{memory loss} that explicitly forces the network to internalize persistent scene context, overcoming sparse supervision occlusion signals.
    
    \item \textbf{Memory Caching \& Stability:} We enable stable, minute-long view synthesis by aggregating historical states via \textbf{memory caching}.
\end{itemize}

\section{Related Work}
\label{sec:related_works}

\noindent\textbf{Novel View Synthesis.} Classical NVS enforces multi-view geometric consistency through explicit 3D representations such as meshes~\cite{hedman2018deep,thies2019deferred,riegler2020free}, multi-plane images~\cite{flynn2016deepstereo,flynn2019deepview,broxton2020immersive,flynn2024quark}, neural radiance fields~\cite{yariv2020multiview,yariv2021volume,wang2021neus,wang2023neus2,mildenhall2021nerf,wang2021ibrnet,lin2022efficient}, point clouds~\cite{aliev2020neural,xu2022point,wang2024dust3r,wang2025vggt}, and 3D Gaussian splatting~\cite{kerbl20233d,huang20242d,wu20244d,xu20254dgt}. 
These representations impose inductive biases that constrain flexibility on scenes deviating from predefined priors, 
motivating more general data-driven architectures~\cite{sitzmann2019scene,sajjadi2022scene}. 
Diffusion-based methods hallucinate unobserved regions~\cite{liu2023zero,watson2022novel,yu2024viewcrafter,shi2023mvdream,liu2023syncdreamer} via 2D generative priors~\cite{ho2020denoising}
but rely on iterative sampling ill-suited to real-time streaming. A separate, representation-free line discards explicit 3D proxy entirely: LVSM~\cite{jin2025lvsm} and LaCT~\cite{zhang2025test_lact} cast NVS as a latent sequence modeling task, tokenizing source views and letting expressive transformers implicitly capture multi-view consistency in a single feed-forward pass. We build on this representation-free formulation for its scalability and freedom from geometric priors. 
Crucially, whereas prior works target offline batch-processing of static scenes, NSTM elevates this paradigm to real-time dynamic streaming via a persistent space-time memory and our decoupled memorization-synthesis architecture.   

\noindent\textbf{Streaming Scene Reconstruction.} %
Unlike offline methods optimizing over an entire sequence~\cite{li2022neural,wu20244d,xu20244k4d,li2024spacetime,yang2024deformable,fridovich2023k,cao2023hexplane},
streaming reconstruction performs causal online updates under strict computational budgets. \emph{Static-scene} methods~\cite{Inovid_Mathias, LiveNVS_LauraFink,OnTheFlyReconstruct_AndreasM} incrementally accumulate explicit representation during online capture but do not model dynamic content. For dynamic scenes, per-frame optimization approaches~\cite{sun20243dgstream,yan2025instant} are too slow for real-time use, while recent work~\cite{wang2025learning} accelerates this using optical flow to reuse prior information, though this local tracking accumulates errors degrading quality over time. We address this by encoding a persistent memory directly into the network's fast weights~\cite{zhang2025test_lact}, retaining context and memorizing scene geometry for temporally consistent synthesis without excessive drift.

\noindent\textbf{Realtime Human Novel View Synthesis.}
Real-time human NVS typically relies on per-subject enrollment~\cite{cao2022authentic,kirschstein2025flexavatar,kirschstein2025avat3r} or parametric body priors~\cite{weng2022humannerf,kocabas2023hugs,li2024animatable}, which prevent zero-shot deployment and often fail on loose garments. 
While real-time non-rigid volumetric fusion~\cite{newcombe2015dynamicfusion,dou2016fusion4d} is prior-free, its non-rigid tracking drifts under fast motion and topological changes. 
Conversely, generalizable feed-forward models~\cite{saito2019pifu,saito2020pifuhd,xiu2022icon,xiu2023econ,zheng2024gpsgaussian,zhou2024gpsgaussianplus} enable zero-shot real-time inference, but remain strictly memoryless and fail in occluded regions. 
NSTM overcomes these limitations: it is enrollment-free and prior-free in both geometry and motion, streaming in real time, zero-shot, while accumulating neural memory as new viewpoints are naturally revealed.

\begin{figure*}[t]
  \vspace{-20pt}
  \centering
  \includegraphics[width=0.98\textwidth]{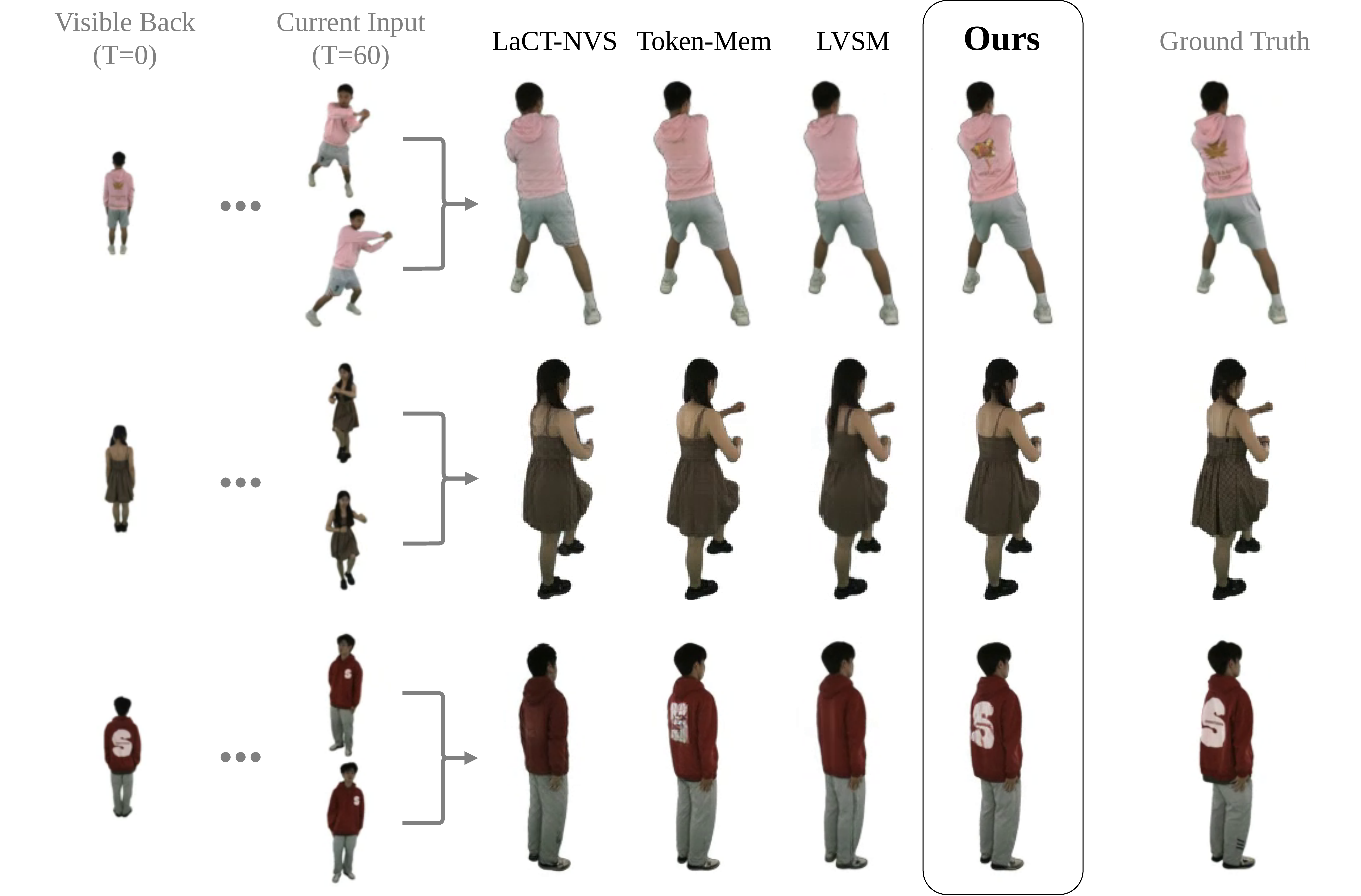} %
   \vspace{-5pt}
   \caption{\textbf{Memory Stress Test at $T=60$}. Despite only observing the back at $T=0$, NSTM accurately recalls the hoodie patterns and hairstyles while fusing with current motion. LVSM makes its best stateless guess in unseen regions, LaCT-NVS suffers from out-of-distribution memory drift, and Token-Mem fails to recall accurate back views. See our supplementary website for the video results.}
  \label{fig:qualitative_results}
  \vspace{-15pt}
\end{figure*}

\noindent\textbf{Memory Modules.} The quadratic cost of Transformers~\cite{vaswani2017attention} on long-context data has driven a broad search for sub-quadratic memory. Linear attention replaces the Softmax operator with a separable kernel~\cite{katharopoulos2020transformers}, compressing context into a fixed-size matrix-valued memory, and a large body of work has since refined the \emph{discipline} of how this memory is written and maintained: through stronger kernels~\cite{kacham2024polysketchformer, arora2024simple}, more expressive learning rules such as the Delta~\cite{schlag2021linear}, Oja~\cite{irie2022neural}, and Omega~\cite{behrouz2025atlaslearningoptimallymemorize} rules, principled forgetting and retention gates~\cite{sun2023retentive, behrouz2026_ALLCONNECTED, hasaniliquid, peng2025rwkv}, and hybrid attention designs~\cite{donghymba, behrouz2025titans}. Because a fixed-size matrix is capacity-limited and underperforms on retrieval, a further line based on Test-Time Memorization/Training~\cite{behrouz2026_ALLCONNECTED, sun2025learninglearntesttime} and Nested Learning~\cite{behrouz2025nested} introduces deep memories whose capacity grows with depth~\cite{behrouz2025titans, behrouz2025atlaslearningoptimallymemorize}, and sub-linearly growing memories via structured matrices or memory caching~\cite{behrouzmemorycaching2026, guo2025log}. We build directly on this memory-discipline lineage. 
Concretely, our memory differs from above in four respects: (1) prior modules are 1-dimensional and sequential, and cannot exploit the rich space-time dependencies in multi-view videos; (2) their coupled update-apply mechanism causes suboptimal information flow along the time dimension, as our ablations confirm (\cref{sec:Ablation}, \cref{tab:ablations}); (3) contrary to these approaches, our design decouples memorization and synthesis by using different modules in our architecture; and (4) we adopt an $L_2$ objective for long-horizon stability; this avoids the magnitude inflation that the dot-product inner loss used across most prior memory modules, including the large-chunk TTT of LaCT~\cite{zhang2025test_lact}, incurs over long rollouts.

\section{Method}
\vspace{-5pt}
\label{sec:method}

We address online novel view synthesis from multi-view dynamic video streams. At each timestep t, we observe N input views $\mathcal{O}_t = \{(\mathbf{I}_{t,i}, \mathbf{P}_{t,i})\}_{i=1}^N$, where $\mathbf{I}_{t,i} \in \mathbb{R}^{H \times W \times 3}$ is an RGB image and $\mathbf{P}_{t,i} \in SE(3)$ is the corresponding camera pose. Because $\mathcal{O}_t$ provides an incomplete observation of the true dynamic scene (e.g., due to occlusion), NSTM synthesizes the target image $\mathbf{I}_{t}^\text{tgt}$ by conditioning the rendering function  on both the current input $\mathcal{O}_t$ and an accumulated implicit memory $\mathcal{W}_{t-1}$:

\begin{equation}
\label{eq:render_eq}
\mathbf{I}_{t}^\text{tgt} = \Phi \left(\mathbf{P}_{t}^\text{tgt}, \mathcal{O}_t; \mathcal{W}_{t-1}\right).
\end{equation}

\noindent Following LVSM~\cite{jin2025lvsm}, we channel-wise concatenate input RGB images with their corresponding per-pixel Pl\"ucker ray maps, using linear layers to patchify them into input tokens. Similarly, we patchify the Pl\"ucker rays of the target camera to form target tokens. These concatenated input and target tokens are then passed through a stack of Transformer blocks, each comprising an attention module, a TTT memory module, and a feed-forward MLP module (\cref{fig:training_regime}). Finally, we extract the processed target tokens and unpatchify them, applying a sigmoid activation to predict the target RGB images (see \supref{sec:detailed_architecture} for  details).

\noindent\textbf{TTT memory module.} Test Time Training/Memorization \cite{sun2025learninglearntesttime, behrouz2026_ALLCONNECTED} defines implicit memory as a set of \emph{fast weights} that are updated at inference time, forming a compressive neural memory. This contrasts with the \emph{slow weights} (i.e., model parameters) that are frozen during inference. In our architecture, we use $\mathcal{W}_{t-1} = \{W_1, W_2, W_3\}$ as the fast weights with the design of a bias-free SwiGLU-MLP:
\begin{equation}
\label{eq:bg_swiglu}
f_\mathcal{W}(x) = \big(\text{SiLU}(W_1 x) \odot (W_3 x)\big)\, W_2,
\end{equation}
Input tokens are projected into keys $k$, queries $q$, and values $v$, then to read from the weights we \textbf{apply} them via ($o = f_\mathcal{W}(q)$). %
The \textbf{update} operation memorizes context via a single gradient step on an inner loss binding keys to values,  
\begin{equation}
\label{eq:bg_update}
\mathcal{W}' = \mathcal{W} - \eta\, \nabla_\mathcal{W}\, \mathcal{L}_{\text{inner}}\big(f_\mathcal{W}(k),\, v\big).
\end{equation}
The inner loss is a design choice that distinguishes different existing architectures and determines the memory management in prioritizing the input tokens. In this paper, we use an $L_2$ inner loss:
\begin{equation}
\label{eq:update}
\mathcal{L}_{\text{inner}} = \| v - f_\mathcal{W}(k) \|_2^2.
\end{equation}
In \cref{tab:ablations,fig:ablations} we show that contrary to the choice of unbounded dot-product loss as in some previous models (such as LaCT~\citep{zhang2025test_lact} or DLA~\citep{behrouz2025atlaslearningoptimallymemorize}) that cause severe magnitude inflation of  $f_\mathcal{W}(k)$ after repeated gradient descent over long horizons, our choice of $L_2$ objective results in long-horizon stability.

\noindent We further stabilize the update by Newton--Schulz orthonormalization of the gradient (Muon~\cite{jordan2024muon}) and $L_2$ weight normalization. Our network is trained end-to-end via an outer-loop task loss over the slow weights, initial fast-weight state, and projections. We denote the fast-weight state at timestep $t$ as $\mathcal{W}_t$
(\cref{subsec:memory_caching}).

\subsection{Decoupled Memorization and Synthesis}
\label{subsec:decoupled_mem}

Na\"ively applying our memory architecture summarized in \cref{sec:method} to live video streams introduces a latency bottleneck. Standard TTT-based models rigidly couple the memory update and apply steps. To achieve high throughput, prior static frameworks~\cite{zhang2025test_lact} process sequence chunks in parallel. However, this relies on \emph{chunk-level buffering}, which introduces unacceptable delays and violates the strict online constraints of live streaming. Conversely, processing the stream sequentially, by forcing a heavy gradient-based memorization at every incoming frame, incurs prohibitive computational overhead of 58ms (see \supref{tbl:perf_breakdown} latency breakdown), severely precluding real-time performance. To overcome this cost, NSTM decouples the frequency of memory updates from novel view synthesis, and decomposes the online inference process into two frequency-asymmetric steps (\cref{fig:training_regime}): 

\noindent\textbf{Periodic Memorization Step:} Rather than updating at every frame, we strictly limit heavy gradient updates to a periodic schedule to compress novel states into the fast-weights.
\noindent\textbf{Per-frame Synthesis Step:} At every timestep $t$, we execute a lightweight query against the persistent memory $\mathcal{W}_{t-1}$ \cref{eq:render_eq}. However, because the scene is dynamic and memory updates are periodic, the memory state and the current observation are inherently misaligned (\cref{fig:training_regime}). To handle this temporal deformation, we adopt a \emph{cross-view attention} mechanism. Jointly attending across the source posed images and the target camera rays explicitly captures ongoing motion. The output then queries the memory module, dynamically fusing current motion with historical context to synthesize the novel view at amortized real-time speeds (\cref{fig:teaser}). Note that this process is different from the per-image attention used in recent TTT-based architectures~\citep{jin2026zipmap, zhang2025test_lact}, which target \emph{static}-scene reconstruction and never model deformation between large-chunk memory writes and small-chunk reads.

\subsection{Disentangled Memory Supervision}
\label{subsec:disentangled_sup}

A na\"ive joint training of the cross-view attention module and the memory module is fundamentally underconstrained. Because disocclusion signals in streaming video are naturally sparse, the model tends to over-rely on attention and bypasses historical memory. To force the persistent internalization of scene context, we introduce an alternating training regime across a sequence of $T$ timesteps. 

\noindent\textbf{Memory Supervision.} As illustrated in \cref{fig:training_regime}, on even timesteps ($t=0, 2, \dots$), we read out from the memory and supervise this readout to reconstruct a random target view. To enforce this loss on the memory alone, it is critical to ensure we read out purely from the memory, completely isolated from the current inputs. Therefore, we introduce additional \emph{memory readout tokens}, representing the camera rays of the target view. Crucially, we restrict the attention mask of the aforementioned cross-view attention module such that these memory tokens only self-attend amongst themselves. These isolated tokens then query the updated memory and pass through the feed-forward layers. Ultimately, they form an image, denoted as $\mathbf{I}_t^{\text{mem}}$, relying entirely on the internalized historical state. We supervise this blind retrieval via an auxiliary \textbf{\emph{memory loss}}:
\begin{equation}
\label{eq:memaux_loss}
\begin{aligned}
    \mathcal{L}_{\text{mem}}^{(t)} &= \mathcal{L}(\mathbf{I}_t^{\text{mem}}, \mathbf{I}_t^{\text{gt}}), \\
    \text{where } \mathcal{L}(\mathbf{I}, \mathbf{I}^{\text{gt}}) &= \text{MSE}(\mathbf{I}, \mathbf{I}^{\text{gt}}) + \lambda_{\text{lpips}} \text{LPIPS}(\mathbf{I}, \mathbf{I}^{\text{gt}}).
\end{aligned}
\end{equation}
$\lambda_{\text{lpips}}=0.5$ is the weight to balance the LPIPS loss~\cite{lpips}.

\noindent\textbf{Synthesis Supervision.} On odd timesteps ($t=1, 3, \dots$), the network operates in standard novel view synthesis mode and we supervise the synthesized target view. Here, the standard target ray tokens freely cross-attend with the tokenized inputs $\mathcal{O}_t$ while querying the persistent memory $\mathcal{W}_{t-1}$ inherited from the immediately preceding memorization step. Consequently, despite dynamic deformations between the current observation and the previous memory state, if a region is occluded, the network learns to dynamically fuse and recover it from this preceding memory. We supervise the output target image from this joint synthesis, denoted as $\mathbf{I}_t^{\text{synth}}$, using the identical loss formulation:

\begin{equation}
\label{eq:synthesis_loss}
\mathcal{L}_{\text{synth}}^{(t)} = \mathcal{L}(\mathbf{I}_t^{\text{synth}}, \mathbf{I}_t^{\text{gt}}).
\end{equation}

\noindent This alternating process continues until all $T$ observations in the sequence are processed. The final training objective minimizes the joint loss across all timesteps, balancing the primary novel view synthesis task against the auxiliary memory regularization:
\begin{equation}
\label{eq:total_loss}
\mathcal{L}_{\text{total}} = \sum_{\substack{t=1 \\ t \text{ is odd}}}^{T} \mathcal{L}_{\text{synth}}^{(t)} + \sum_{\substack{t=0 \\ t \text{ is even}}}^{T} \mathcal{L}_{\text{mem}}^{(t)},
\end{equation}

\noindent\textbf{Training-Inference Asymmetry.} While we train with a 1:1 alternating ratio, our framework can execute significantly more synthesis steps than memory updates during inference. This discrepancy does not degrade performance because the synthesis steps between two memorization steps read out from the same frozen memory state and are completely independent of one another. See our supplementary website where we synthesize 29 frames for every 1 step of memorization (29:1).

\subsection{Memory Caching}
\label{subsec:memory_caching}
\begin{figure}[t]
    \vspace{-5pt}
    \centering
    \includegraphics[width=0.5\textwidth]{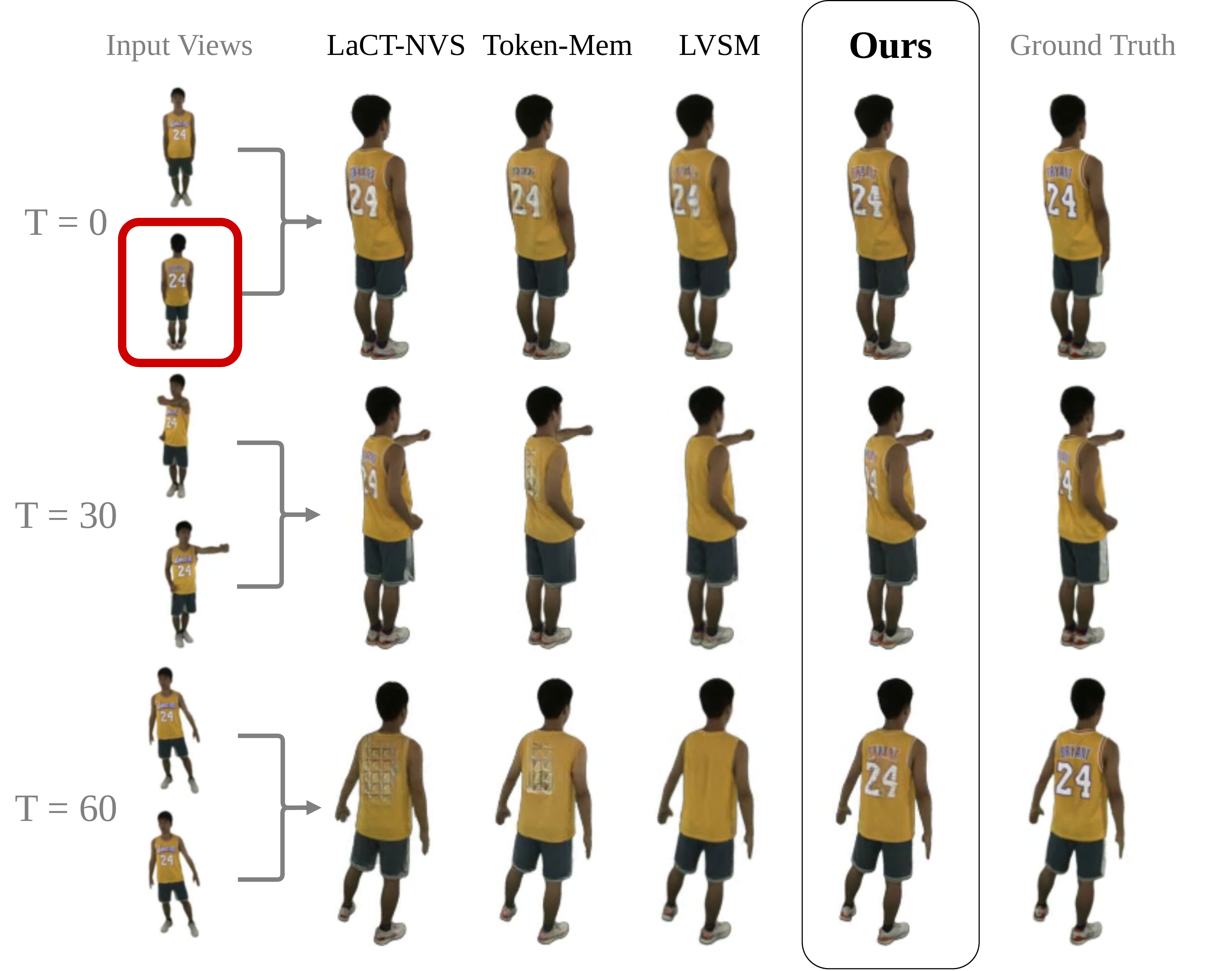} %
    \caption{\textbf{Memory Stress Test Over Time.} We show the back view at T=0. For subsequent timestamps, we show only frontal views and task the model with synthesizing the occluded back. NSTM maintains high-fidelity recall over time while baseline models suffer from  drift or collapse to stateless prior over time.}
    \label{fig:stability_vert}
    \vspace{-20pt}
\end{figure}

TTT suffers from RNN drift over long-horizon updates. We observe this as memory decay, stemming from optimization instability, in the LaCT-NVS~\cite{zhang2025test_lact} column of \cref{fig:stability_vert}. To address this problem, we adapt a memory caching strategy recently proposed for large language models~\cite{behrouzmemorycaching2026, behrouz2026_ALLCONNECTED}.

Concretely, the sequential application of our periodic memory update generates a sequence of evolving memory states $\mathcal{W}_0, \mathcal{W}_1, \mathcal{W}_2, \dots$, where the ones between updates are the same. Note, at any timestep $t$, the active memory is $\mathcal{W}_{t-1}$ since the memory is not reset after a snapshot is taken. We average over the active memory and every $K$ past memory states, which can be implemented recursively. 
\begin{gather*}
\label{eq:cache_main}
\bar{\mathcal{W}}_{0} = \mathcal{W}_{0},\  \bar{\mathcal{W}}_{n} = \frac{n-1}{n}\bar{\mathcal{W}}_{n-1} + \frac{1}{n}\mathcal{W}_{nK}, \forall n>1,\\
\hat{\mathcal{W}}_{t-1} = \frac{1}{m + 1} \left( \mathcal{W}_{t-1} + \sum_{i=1}^{m} \mathcal{W}_{iK} \right) = \frac{\mathcal{W}_{t-1} + m\bar{\mathcal{W}}_m}{m+1}
\end{gather*}
where $m = \lfloor (t-1)/K \rfloor$. $\bar{\mathcal{W}}_m$ is the average of every $K$ past states. In practice, we only keep one extra set of weights $\bar{\mathcal{W}}_m$ to track the whole average. This aggregated state $\hat{\mathcal{W}}_{t-1}$ replaces standard memory conditioning (Eq.~\ref{eq:render_eq}), stabilizing the parameter space and mitigating the out-of-distribution drift caused by continuous gradient updates.
Importantly, this memory caching strategy fails on standard, coupled TTT architectures. Its effectiveness fundamentally relies on our cross-view attention to explicitly handle spatial deformations, as we empirically validate in \cref{sec:Ablation}.

\section{Experiments}
\label{sec:experiments}
We validate NSTM's ability to maintain real-time, high-fidelity, minute-long memory under severe occlusions. We detail our experimental setup in \cref{subsec:setup}, evaluate memory persistence and synthesis fidelity in \cref{subsec:results}, deconstruct our architecture via an ablation study in \cref{sec:Ablation}, and analyze our amortized efficiency in \cref{subsec:efficiency}. We encourage readers to view our supplementary website for convincing comparisons.

\subsection{Experimental Setup}
\label{subsec:setup}

\textbf{Dataset.} We evaluate on MVHumanNet++~\cite{li2025mvhumannet++}, featuring 60,000 multi-view motion sequences across 4,500 identities and 9,000 outfits, using a 90\%/10\% train/test split. The dataset's wide view coverage, unconstrained motion and minute-long sequence length make it an ideal testbed for dynamic NVS and long-term memory tasks. 

\noindent\textbf{Evaluation Protocol.} MVHumanNet++~\cite{li2025mvhumannet++} provides annotated train/eval sets of 60-frame sequences (subsampled at 1.2 FPS from 50s of raw video), enabling minute-scale evaluation. However, measuring long-term memorization in the wild is challenging as natural disocclusions are spatially and temporally sparse -- only 17 of 390 test scenes exhibit full rotations that can best test memorization. To comprehensively evaluate memory, we use two protocols. (1) \emph{Memory Stress Test~(\cref{subsec:results}):} A forced-occlusion setup to evaluate memorization across all scenes. The model observes full front and back views at $T=0$. For subsequent frames, we restrict inputs to frontal stereo views and task the model with synthesizing the occluded back. We evaluate at the final frame (T = 60), quantifying memory persistence at the minute-scale limit. (2) \emph{Memory from Natural Rotations~(\cref{subsec:main_natural_memory_test_results}):} For the 17 full-rotation scenes %
we provide fixed frontal stereo inputs and evaluate synthesis as the back naturally reveals and occludes. To avoid the backgrounds dominating  metrics, we report masked mPSNR and mSSIM isolated over the dynamic foreground. All evaluations are done at $256\times 256$ resolution on one H100. 

\noindent\textbf{Baselines.} We compare against one stateless and two stateful baselines. \textbf{LVSM}~\cite{jin2025lvsm} uses its exact original architecture, trained with the same $L_2$+LPIPS objective and learning-rate schedule as all other models, with training camera coordinates relativized to each frame's target camera. \textbf{LaCT-NVS}~\cite{zhang2025test_lact} retains the original LaCT architecture and its dot-product inner loss, adapted to causal per-frame updates for streaming (a variant equipped with our $L_2$ inner loss is analyzed in \cref{tab:ablations}). 
\textbf{Token-Mem} adapts the recurrent token-state memory of CUT3R~\cite{cut3r} ($768$ tokens $\times$ $768$ channels) to posed NVS: unlike the pose-free original, input and target views are conditioned on Pl\"ucker ray embeddings using the same tokenization as LVSM and NSTM; it is trained with the same $L_2$+LPIPS objective. Our memory tests and ablation study follow the original designs of the stateful baselines, updating their memory every frame, while NSTM updates every two frames via our memorization-synthesis decoupling; we study the effect of evaluating all methods under the same update schedule in \supref{sec:sup_update_interval_analysis}. All stateful models share identical training data, curriculum, optimizer, and reconstruction losses, with camera coordinates relativized to the first target camera. Full training details are provided in \supref{sec:supp_implementation}.

\noindent\textbf{Implementation details.} NSTM and all baselines are implemented in JAX and trained with identical optimizer hyperparameters and a batch size of 128. All $256 \times 256$ stateful models adopt a two-stage training curriculum: memory-bootstrap pretraining on 4-frame clips (64 H100 GPUs), followed by fine-tuning on 24-frame clips (128 H100 GPUs) for free-viewpoint synthesis and long-term memory. 
The stateless LVSM~\cite{jin2025lvsm} trains in a single stage on the final frame of the same 4-frame clips, ensuring an identical data distribution. See \supref{sec:supp_implementation} for full details. Inference latency is evaluated on a single H100 GPU at $256 \times 256$. We set $\lambda_{\text{LPIPS}} = 0.5$.

\begin{table}[t]
\centering
\resizebox{\linewidth}{!}{%
\begin{tabular}{lcccccc}
\toprule
Method & PSNR$\uparrow$ & SSIM$\uparrow$ & LPIPS$\downarrow$ & DISTS$\downarrow$ & mPSNR$\uparrow$ & mSSIM$\uparrow$ \\
\midrule
LVSM & \cellcolor{second}28.99 & \cellcolor{second}0.9662 & \cellcolor{second}0.0257 & \cellcolor{second}0.1034 & \cellcolor{second}19.34 & \cellcolor{third}0.5670 \\
LaCT-NVS & 25.85 & 0.9500 & 0.0360 & 0.1236 & 16.25 & 0.5158 \\
Token-Mem & \cellcolor{third}28.97 & \cellcolor{third}0.9661 & \cellcolor{third}0.0253 & \cellcolor{third}0.1019 & \cellcolor{third}19.85 & \cellcolor{second}0.6220 \\
Ours & \cellcolor{best}29.99 & \cellcolor{best}0.9701 & \cellcolor{best}0.0225 & \cellcolor{best}0.0919 & \cellcolor{best}20.71 & \cellcolor{best}0.6286 \\
\bottomrule
\end{tabular}
}
\vspace{-5pt}
\caption{
\textbf{Long-horizon Memory Stress Test~(\cref{subsec:results})}, measured on last frame $T{=}60$. NSTM ranks best on every metric, leading LaCT-NVS by 4.46 dB and Token-Mem by 0.86 dB. As a stateless model, LVSM fails to reconstruct details in the back view when only observing frontal views as seen in~\cref{fig:qualitative_results} and~\cref{fig:stability_vert}. The \colorbox{best}{best}, \colorbox{second}{second-best}, and \colorbox{third}{third-best} results are highlighted. 
}
\label{tbl:main_results}
\vspace{-15pt}
\end{table}

\subsection{Memory Tests}
\label{subsec:results}
\begin{figure*}[t]
\vspace{-20pt}
  \centering
  \includegraphics[width=\textwidth]{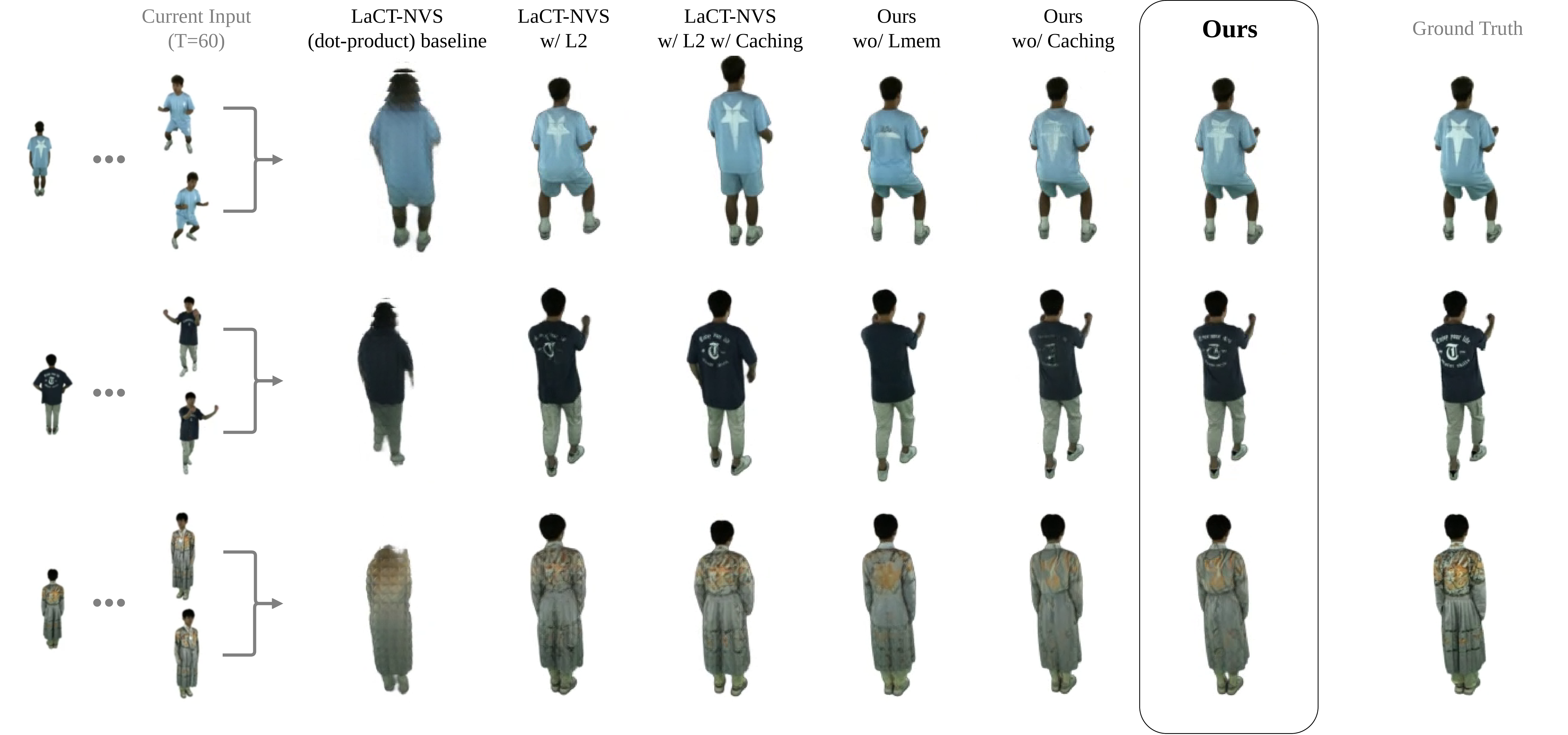}
    \vspace{-20pt}
    \caption{
    \textbf{Qualitative ablation study.} At $T=60$, each model synthesizes the back view that has been occluded after $T=0$. \textbf{LaCT-NVS} demonstrates the instability of the inner dot-product loss. In contrast, \textbf{LaCT-NVS w/ $L_2$} greatly improves stability through our $L_2$ inner loss choice, though it still suffers from some drift. \textbf{LaCT-NVS w/ $L_2$ w/ Caching} freezes into an averaged past pose, demonstrating that our memorization-synthesis decoupling with cross-view attention is essential for effective memory caching. Dropping $\mathcal{L}_{\text{mem}}$ from our method collapses recall, validating our disentangled memory supervision design (\cref{subsec:disentangled_sup}). Furthermore, dropping memory caching reduces the recall horizon. Only the full model (\textbf{Ours}) provides minute-scale recall of the historical back view while capturing current motion. See our supplementary website for the video results.
    }
    \label{fig:ablations}
    \vspace{-15pt}
\end{figure*}

\noindent\textbf{Memory Stress Test.} We present our forced-occlusion protocol quantitative comparisons in \cref{tbl:main_results} and qualitative visual evidence in \cref{fig:qualitative_results}. NSTM ranks best on every metric at minute-long horizon ($T=60$). Prior stateful models exhibit their fragility: Token-Mem forgets historical context, its performance comparable to the stateless LVSM that never observed the back view. LaCT-NVS trails NSTM by 4.46 dB mPSNR due to the instability of its continuous update mechanism. This quantitative collapse is clearly visible in \cref{fig:qualitative_results} and \cref{fig:stability_vert}: where LaCT-NVS's memory drifts significantly, while LVSM produces blurry outputs in unseen regions (e.g. the back) due to its stateless nature. On the other hand, NSTM maintains highly robust long-term memory which is especially pronounced in our masked foreground metrics (mPSNR/mSSIM), proving the model accurately locks in the dynamic subject's identity rather than relying on the background. Qualitatively (\cref{fig:qualitative_results}), NSTM faithfully reconstructs specific, temporarily occluded details---such as complex logo and hair styles---long after they disappear from the input stream. This sustained high-fidelity recall directly validates the stabilizing effect of our $L_2$ objective and memory caching strategy. 

\begin{table}[t]
\centering
\resizebox{\linewidth}{!}{%
\begin{tabular}{lcccccc}
\toprule
Method & PSNR$\uparrow$ & SSIM$\uparrow$ & LPIPS$\downarrow$ & DISTS$\downarrow$ & mPSNR$\uparrow$ & mSSIM$\uparrow$ \\
\midrule
LVSM & \cellcolor{second}28.81 & \cellcolor{third}0.9656 & \cellcolor{third}0.0296 & \cellcolor{third}0.1298 & \cellcolor{second}19.69 & \cellcolor{third}0.5272 \\
LaCT-NVS & 27.01 & 0.9551 & 0.0340 & 0.1414 & 17.93 & 0.4982 \\
Token-Mem & \cellcolor{third}28.27 & \cellcolor{second}0.9626 & \cellcolor{second}0.0293 & \cellcolor{second}0.1250 & \cellcolor{third}19.45 & \cellcolor{second}0.5413 \\
Ours & \cellcolor{best}29.68 & \cellcolor{best}0.9691 & \cellcolor{best}0.0250 & \cellcolor{best}0.1107 & \cellcolor{best}20.88 & \cellcolor{best}0.5997 \\
\bottomrule
\end{tabular}
}
\vspace{-5pt}
\caption{
\textbf{Long-horizon Memory from Natural Rotations~(\cref{subsec:main_natural_memory_test_results})}, evaluated at T=60. NSTM ranks best on every metric, leading LaCT-NVS by 2.95 dB and Token-Mem by 1.43 dB in mPSNR. 
    See~\supref{sec:supp_natural_memory_test} for further qualitative and quantitative results and our supplementary website for video results.
}
\label{tbl:natural_lastframe_results}
\vspace{-5pt}
\end{table}

\label{subsec:main_natural_memory_test_results}
\noindent\textbf{Memory from Natural Rotations.} We evaluate the benefit of memory in natural video sequences in \cref{tbl:natural_lastframe_results}.
The strength of NSTM remains evident, outperforming Token-Mem and LaCT-NVS by  1.43 dB and 2.95 dB mPSNR on minute-long sequences in the 17 full-rotation scenes. The stateless LVSM is unable to memorize content revealed by these natural disocclusions.

\subsection{Ablation Study}
\label{sec:Ablation}
\begin{table}[t]
\centering
\resizebox{0.5\textwidth}{!}{%
\begin{tabular}{lccccc}
\toprule
 & \multicolumn{5}{c}{Ablation Study (over 60 timesteps)} \\
 \cmidrule(lr){2-6}
Method & PSNR$\uparrow$ & SSIM$\uparrow$ & LPIPS$\downarrow$ & mPSNR$\uparrow$ & mSSIM$\uparrow$ \\
\midrule
LaCT-NVS (dot-product) baseline & 22.32 & 0.9275 & 0.0616 & 13.49 & 0.4942 \\
LaCT-NVS w/ $L_2$ & 27.77 & 0.9594 & 0.0275 & 19.08 & 0.6189 \\
LaCT-NVS w/ $L_2$ w/ Mem Caching & 20.25 & 0.9135 & 0.0688 & 12.03 & 0.4707 \\
Ours wo/ $\mathcal{L}_{\text{mem}}$ & \cellcolor{third}28.23 & \cellcolor{third}0.9639 & \cellcolor{third}0.0251 & \cellcolor{third}19.36 & \cellcolor{third}0.6198 \\
Ours wo/ Mem Caching & \cellcolor{second}29.24 & \cellcolor{second}0.9673 & \cellcolor{second}0.0232 & \cellcolor{second}19.78 & \cellcolor{second}0.6321 \\
Ours w/ Mem Caching & \cellcolor{best}30.09 & \cellcolor{best}0.9705 & \cellcolor{best}0.0217 & \cellcolor{best}20.79 & \cellcolor{best}0.6537 \\
\bottomrule
\end{tabular}
}
\caption{\textbf{Quantitative Ablation Study}. We ablate all variants under stage 1 pretraining before finetuning. Metrics are averaged over 60 timesteps over 390 MVHumanNet++~\cite{li2025mvhumannet++} eval scenes.}
\label{tab:ablations}
\vspace{-15pt}
\end{table}

We validate our design choices in \cref{tab:ablations} and \cref{fig:ablations}, measuring the impact of each component on the pretrained models using the 60-frame memory stress test protocol.

\noindent\textbf{Stability with $L_2$  Objective.} We first establish a naïve causal baseline (LaCT-NVS dot-product)~\cite{zhang2025test_lact} that updates memory at every frame without decoupling memorization from synthesis. As observed, the unbounded dot-product inner loss causes magnitude inflation over continuous updates, resulting in poor performance (22.32 PSNR). We next replace the dot-product loss with a stable $L_2$ objective (LaCT-NVS w/ $L_2$), which also represents a no-decoupling variant of our architecture. $L_2$ significantly improves stability and synthesis quality (+5.45 PSNR).

\noindent\textbf{Memory Caching Requires Decoupled Training.} While memory caching is designed for long-term stability, it fundamentally fails on coupled architectures. Row 3 evaluates the LaCT-NVS with $L_2$ baseline by performing its standard update-and-apply step at every frame, but uniquely conditioning it on the aggregated memory cache (\cref{subsec:memory_caching}). This causes a severe performance drop (20.25 PSNR). The synthesized output becomes visually ``stuck'' in a historical pose (see LaCT-NVS w/ $L_2$ w/ caching column in \cref{fig:ablations}). This occurs because temporal motion is inherently baked into the fast-weights; aggregating these weights creates a superposition of past poses. In our decoupled architecture (\cref{subsec:decoupled_mem}), the cross-view attention explicitly learns to resolve and deform from this superimposed historical memory to match the current observation. The tightly coupled baseline lacks this explicit deformation mechanism, causing it to simply output the averaged past poses. This proves that structurally decoupling memorization from synthesis is necessary to successfully aggregate historical memory.

\noindent\textbf{Disentangled Supervision Enhances Memory Persistence.} When we fully decouple the architecture but remove the auxiliary memory loss $\mathcal{L}_{\text{mem}}$ (Ours wo/ $\mathcal{L}_{\text{mem}}$), performance drops to 28.23 PSNR. Without explicit pressure to reconstruct the scene independently, the network takes a shortcut, over-relying on the cross-view attention's current visual context and partially collapsing to a stateless prior. Introducing $\mathcal{L}_{\text{mem}}$ forces robust geometric internalization (+1.01 PSNR).

\noindent\textbf{Long-term Stability via Memory Caching.} Finally, applying our memory caching strategy to the fully decoupled and properly supervised model (Ours w/ Mem Caching) yields our highest performance (30.09 PSNR), successfully mitigating long-term weight drift and preserving minute-long context.

\subsection{Real-Time Inference Efficiency}
\label{subsec:efficiency}
To validate our claim of real-time, long-horizon generation, we compare the amortized inference latency of NSTM against \emph{Temporal-LVSM}, an adaptation of LVSM~\cite{jin2025lvsm} for dynamic video sequences. To handle long contexts in this baseline, we compute dense attention across all past observations that we want to memorize. As shown in \cref{fig:teaser}, this imposes a quadratic computational cost, causing latency to spike and precluding real-time rendering on continuous streams. In contrast, NSTM architecturally bypasses this bottleneck, achieving constant inference latency regardless of video length. Our decoupling strategy breaks the tie between temporal horizon and computational cost, unlocking sustained, real-time dynamic view synthesis.

\vspace{-8pt}
\section{Conclusion}
\label{sec:limitations}
We presented Neural Space-Time Memory (NSTM), a novel online dynamic NVS framework operating on multi-view streaming videos. By disentangling periodic memory updates from per-frame transformer-based deformation modeling and introducing a targeted memory auxiliary loss, NSTM internalizes persistent scene context while maintaining amortized real-time speed. Furthermore, our memory caching strategy and stable $L_{2}$ update rule unlock minute-long memory retention, achieving state-of-the-art %
long-horizon memory persistence. While NSTM provides a principled solution for streaming dynamic NVS, we discuss our limitations and avenues for future research in~\supref{sec:supp_limitations}.

    \section{Acknowledgements}
We would like to thank Michael Broxton and Ryan Overbeck for their encouragement and support of the project at Google. We also thank John Flynn and Osman Ulusoy for their valuable discussions and feedback on the manuscript, and Tianyuan Zhang and Bowei Chen for early clarifying discussions on LaCT. Finally, we extend our appreciation to Xiaoguang Han, Hongjie Liao, and Yihao Zhi for providing full-length MVHumanNet++ videos for our experiments. %

{
    \small
    \bibliographystyle{ieeenat_fullname}
    \bibliography{main}
}
    \appendix

    \section{Supplementary Material}
    Our Supplementary Material presents additional experiments on memory persistence, general novel-view synthesis, and generalization beyond human-centric scenes via a category-agnostic object-centric NVS test in~\cref{sec:sup_additional_experiments}. We provide full implementation, training, inference and model architecture details in~\cref{sec:supp_implementation}. Finally, we discuss limitations and avenues for future research in~\cref{sec:supp_limitations} and concurrent work in~\cref{sec:supp_concurrent}.

\section{Additional Experiments}
\label{sec:sup_additional_experiments}

\subsection{Memory from Natural Rotations Test} 
\label{sec:supp_natural_memory_test}

To evaluate NSTM on the general setting of natural video sequences where occlusions can happen naturally at any point, we evaluate 17 scenes in MVHumanNet++ that feature back reveal during natural rotations. This is in contrast to our ``Memory Stress Test'' (\mainref{tbl:main_results}) where the model sees the back view only at timestep 0 and needs to remember it while receiving only frontal views over the entire duration of the video. To test, we provide fixed frontal stereo inputs (\texttt{cam15} and \texttt{cam01}) and evaluate synthesis as the back naturally reveals and occludes.

\begin{figure*}[t]
  \vspace{-5pt}
  \centering
  \includegraphics[width=\textwidth]{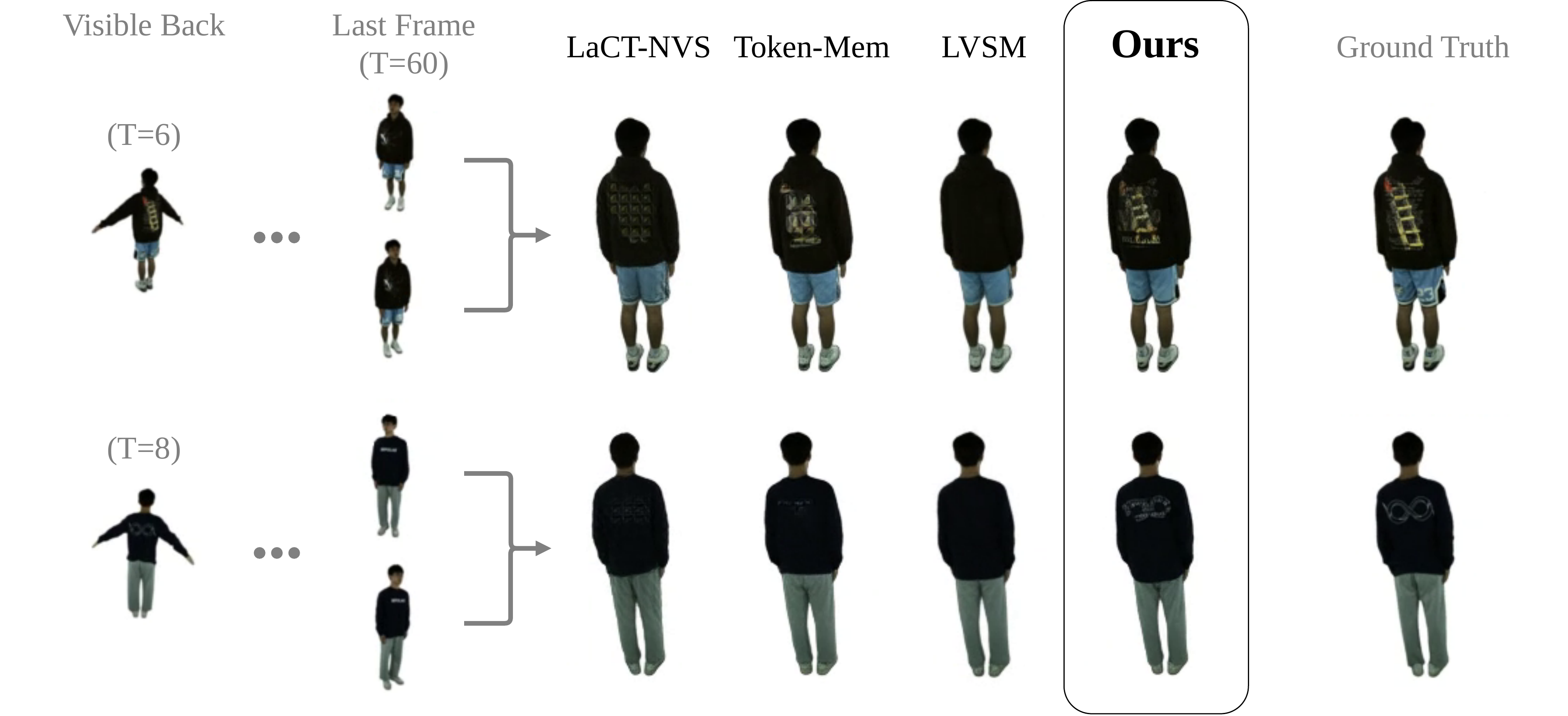} %
  \caption{\textbf{Memory from Natural Rotations Qualitative Results}. 
  We evaluate our method against the baselines on sequences featuring natural rotations that reveal the back of the subject, which the models must synthesize.
  Left column shows at which timestep the back is revealed. NSTM recalls the clothing graphics. The stateless LVSM cannot synthesize the unseen graphics, while LaCT-NVS and Token-Mem show a degraded memory. See supplementary website for more video results. }
  \label{fig:natural_memory}
  \vspace{-5pt}
\end{figure*}

\noindent\textbf{Results and Discussions.} We present additional quantitative comparisons in \cref{tbl:natural_memory_results} and qualitative visual evidence in \cref{fig:natural_memory} measuring performance both averaged across all frames and on the final frame in the $(t=60)$ sequence. The strength of NSTM remains evident, outperforming Token-Mem and LaCT-NVS~\cite{zhang2025test_lact} by  1.43 dB and 2.95 dB mPSNR on last frame measurement. Since LVSM~\cite{jin2025lvsm} is a stateless model the average and last frame performance are almost identical, demonstrating that it is not able to remember past details that would allow it to increase its accuracy. 

\subsection{General Novel View Synthesis Test}
\label{sec:nvs}
In addition to the long-horizon memory performance shown in the memory stress test in \mainref{subsec:results} and natural memory test in~\cref{sec:supp_natural_memory_test}, NSTM is fundamentally designed to be a highly versatile dynamic NVS framework. To evaluate its generalized synthesis capabilities, we conduct experiments under a standard, unconstrained streaming protocol. 

\noindent\textbf{Evaluation Protocol.} To simulate typical real-world telepresence or broadcasting scenarios where input camera rigs are fixed, we restrict the streaming inputs to only two static frontal views (\texttt{cam15} and \texttt{cam01}). The model is then tasked with rendering the dynamic scene from arbitrary time-varying target viewpoints. 
Because disocclusions under this protocol are spatially and temporally sparse, the benefit of long-term memory is naturally limited. Thus, we use this setup strictly to isolate and stress-test the model's spatial generalization -- specifically, its ability to synthesize arbitrary unseen viewpoints from instantaneous inputs.

\noindent\textbf{Results and Discussions.} 
Despite temporal context providing limited benefits, Table~\ref{tbl:general_nvs} shows that NSTM consistently outperforms the stateful LaCT-NVS and Token-Mem baselines. Furthermore, NSTM remains competitive in standard structural metrics (PSNR, SSIM, and mPSNR) against the purely stateless LVSM baseline, while outperforming  on LPIPS, DISTS, and mSSIM~\cite{DBLP:journals/corr/abs-2004-07728_DISTS}.
This demonstrates our model's spatial rendering capabilities.

\begin{table}[t]
    \centering
    \begin{tabular}{lcc}
      \toprule
      Step & Latency (ms) $\downarrow$ & FPS $\uparrow$ \\
      \midrule
      Memorization & 58.14 & 17.19 \\
      Synthesis    & 27.01 & 37.02 \\
      \bottomrule
    \end{tabular}
  \caption{\textbf{Computation Asymmetry.} The memorization step (including both memory update and apply) is significantly slower than the synthesis step (apply only). Inference performance breakdown measured on a single H100 GPU with a resolution of 256x256.}
  \label{tbl:perf_breakdown}
  \vspace{-15pt}
\end{table}

\begin{table*}[t]
\vspace{-5pt}
  \centering
  \resizebox{\textwidth}{!}{%
  \begin{tabular}{lcccccccccccccc}
  \toprule
   & \multicolumn{6}{c}{Average over T=60} &  & \multicolumn{6}{c}{Last frame} \\
   \cmidrule(lr){2-7} \cmidrule(lr){9-14}
  Method & PSNR$\uparrow$ & SSIM$\uparrow$ & LPIPS$\downarrow$ & DISTS$\downarrow$ & mPSNR$\uparrow$ & mSSIM$\uparrow$ &  & PSNR$\uparrow$ &
  SSIM$\uparrow$ & LPIPS$\downarrow$ & DISTS$\downarrow$ & mPSNR$\uparrow$ & mSSIM$\uparrow$ \\
  \midrule
  LVSM & \cellcolor{third}28.81 & \cellcolor{third}0.9654 & \cellcolor{third}0.0286 & 0.1216 & \cellcolor{third}19.21 & 0.5069 &  & \cellcolor{second}28.81 & \cellcolor{third}0.9656 & \cellcolor{third}0.0296 & \cellcolor{third}0.1298 &
  \cellcolor{second}19.69 & \cellcolor{third}0.5272 \\
  LaCT-NVS & \cellcolor{second}29.13 & \cellcolor{second}0.9667 & \cellcolor{second}0.0253 & \cellcolor{best}0.1045 & \cellcolor{second}19.98
  & \cellcolor{second}0.5850 &  & 27.01 & 0.9551 & 0.0340 & 0.1414 & 17.93 & 0.4982 \\
  Token-Mem & 27.76 & 0.9607 & 0.0299 & \cellcolor{third}0.1198 & 18.69 & \cellcolor{third}0.5254 &  & \cellcolor{third} 28.27 & \cellcolor{second}0.9626 & \cellcolor{second}0.0293 & \cellcolor{second}0.1250 & \cellcolor{third}19.45 & \cellcolor{second}0.5413 \\
  
  Ours & \cellcolor{best}29.53 & \cellcolor{best}0.9692 & \cellcolor{best}0.0247 & \cellcolor{second}0.1056 & \cellcolor{best}20.39 &
  \cellcolor{best}0.5928 &  & \cellcolor{best}29.68 & \cellcolor{best}0.9691 & \cellcolor{best}0.0250 & \cellcolor{best}0.1107 &
  \cellcolor{best}20.88 & \cellcolor{best}0.5997 \\
  \bottomrule
  \end{tabular}
  }
    \caption{\textbf{Memory from Natural Rotations Test.} \colorbox{best}{Best}, \colorbox{second}{second best}, and \colorbox{third}{third best} are highlighted. Memory is critical for this evaluation protocol: the sequences feature natural rotations that reveal the back of the subject, which the models must accurately synthesize. Because the memory of stateful baselines degrades over time, NSTM significantly outperforms Token-Mem and LaCT-NVS by 1.43 dB and 2.95 dB in mPSNR, respectively, when measured on the final frame. Furthermore, by also surpassing the stateless LVSM, NSTM clearly demonstrates the advantage of our robust memory persistence. See our supplementary website for video results.}
  \label{tbl:natural_memory_results}
  \end{table*}

\begin{table}[t]
\vspace{-5pt}
\centering
\label{tab:main_results}
\resizebox{0.5\textwidth}{!}{%
\begin{tabular}{lcccccc}
\toprule
 & \multicolumn{6}{c}{Novel View Synthesis Over 60 Timesteps} \\
 \cmidrule(lr){2-7}
Method & PSNR$\uparrow$ & SSIM$\uparrow$ & LPIPS$\downarrow$ & DISTS$\downarrow$ & mPSNR$\uparrow$ & mSSIM$\uparrow$ \\
\midrule
LVSM & \cellcolor{best}29.48 & \cellcolor{best}0.9699 & \cellcolor{second}0.0225 & \cellcolor{second}0.0917 & \cellcolor{best}20.10 & \cellcolor{second}0.6509 \\
LaCT-NVS & \cellcolor{third}26.84 & \cellcolor{third}0.9567 & \cellcolor{third}0.0281 & \cellcolor{third}0.0997 & \cellcolor{third}17.62 & \cellcolor{third}0.5869 \\
Token-Mem & 25.33 & 0.9470 & 0.0328 & 0.1056 & 16.50 & 0.5692 \\
Ours & \cellcolor{second}29.11 & \cellcolor{second}0.9696 & \cellcolor{best}0.0223 & \cellcolor{best}0.0894 & \cellcolor{second}19.93 & \cellcolor{best}0.6703 \\
\bottomrule
\end{tabular}
}
\caption{\textbf{General Novel View Synthesis Comparison.} \colorbox{best}{best}, \colorbox{second}{second best}, and \colorbox{third}{third best} are highlighted. 
Given stereo inputs from fixed frontal cameras, all models synthesize arbitrary unseen viewpoints over time. The metrics are averaged over 60 timesteps across 390 evaluation scenes. NSTM significantly outperforms prior stateful baselines (LaCT-NVS and Token-Mem) by successfully mitigating long-term memory degradation. Because the sparsity of disocclusions in this protocol limits the benefits of historical context, NSTM performs on par with the stateless LVSM baseline, confirming its strong general NVS capabilities. See videos in our supplementary website for qualitative visual results.}
\label{tbl:general_nvs}
    \vspace{-17pt}
\end{table}

\subsection{Inference Speed}
As mentioned in the Introduction, the memory update and apply operations exhibit asymmetric computational costs. As detailed in \cref{tbl:perf_breakdown}, a full memorization step (both update and apply) requires 58.14 ms, whereas a synthesis step (apply only) takes just 27.01 ms. Consequently, by scheduling memorization at 1 FPS and synthesis at 30 FPS, our framework achieves an amortized rendering speed of 28.1 ms per frame (see~\mainref{fig:teaser}). All measurements were conducted at $256 \times 256$ resolution on a single NVIDIA H100 GPU.

\subsection{Update Interval Analysis}
\label{sec:sup_update_interval_analysis}
To study the effects of update-frequency across baselines we evaluate all stateful methods under identical memory-update intervals $\tau \in \{1,2,4,8,16,32\}$ using the same evaluation protocol as our Memory from Natural Rotations test (\cref{sec:supp_natural_memory_test}). At each $\tau$, all methods perform their memory update on the same frames, equalizing the update interval across methods and, for the gradient-based memories (NSTM, LaCT-NVS), the number of inference-time inner-loss gradient steps.

\begin{figure}[t]
\vspace{-5pt}
  \centering
  \begin{subfigure}[b]{0.485\linewidth}
    \includegraphics[width=\linewidth]{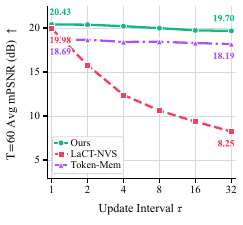} %
  \end{subfigure}\hfill
  \begin{subfigure}[b]{0.485\linewidth}
    \includegraphics[width=\linewidth]{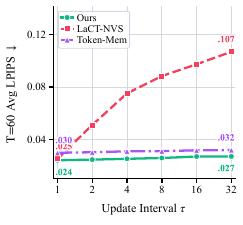} %
  \end{subfigure}
  \vspace{-4pt}
  \caption{\textbf{Update Interval Analysis}
  ($\mathrm{T}{=}60$
  sequence average over 17 Natural Rotation scenes). Our method is near stride-invariant, losing only
  $0.73$\,dB mPSNR ($20.43\rightarrow19.70$) and $0.003$ LPIPS
  ($.024\rightarrow.027$), whereas LaCT-NVS degrades by
  $11.73$\,dB mPSNR ($19.98\rightarrow8.25$) and $4.2\times$ in LPIPS   ($.025\rightarrow.107$).  Token-Mem is stable but capacity-limited, trailing ours at every interval.}
  \label{fig:update-analysis-tau-pair}
  \vspace{-5pt}
\end{figure}

\noindent\textbf{Results and Discussions.} As shown in~\cref{fig:update-analysis-tau-pair} NSTM is near-invariant to $\tau$ and outperforms both baselines at every matched interval — by $4.61$/$1.72$\,dB mPSNR over LaCT-NVS/Token-Mem at $\tau{=}2$, and by $0.44$\,dB over LaCT-NVS even at its native $\tau{=}1$. As the update interval increases, LaCT-NVS collapses ($19.98\rightarrow8.25$\,dB) as it lacks a deformation mechanism to reconcile stale memory with current motion. Token-Mem is $\tau$-insensitive but capacity-limited ($\approx$$18$\,dB).

\subsection{Category-Agnostic Object-Centric NVS Test}
\label{subsec:sup:sec:object_centric_nvs}
Thus far, we have evaluated NSTM on human-centric scenes from MVHumanNet++~\cite{li2025mvhumannet++}. While generic large-scale long-horizon multi-view dynamic scene data is limited, we aim to demonstrate that our framework is prior-free and generalizes beyond just humans. Therefore, we train and evaluate all methods from scratch—without any MVHumanNet++ pretraining—on a fully dynamic object-centric dataset generated with Kubric~\cite{greff2021kubric}.
Objects are drawn from all 55 categories of ShapeNetCore.v2~\cite{shapenet2015}, and a single model is trained across them; validation scenes are built exclusively from a held-out 10\% of the assets, so no evaluation object is ever seen during training. We design the scenes to test memory under sustained motion and occlusion. A primary object fixed at the center rotates continuously, exposing different facets to the cameras over the 30-second sequence, while a smaller secondary object from the same asset pool is launched with a random velocity in zero gravity, bouncing between the primary object and the walls of an invisible 3\,m bounding box. Scenes are rendered against a white background at $256 \times 256$ resolution. All cameras lie on a sphere of radius 3\,m looking at the scene center. Each scene has three randomly sampled cameras, fixed over the sequence: a narrow-baseline stereo input pair and a target view well outside that baseline. The training split contains 16{,}384 scenes at 6 FPS (180 frames each) and the validation split 100 scenes at 30 FPS (900 frames each).
\begin{table}[t]
\vspace{-5pt}
\centering
\resizebox{\linewidth}{!}{%
\begin{tabular}{lcccc}
\toprule
Method & PSNR$\uparrow$ & SSIM$\uparrow$ & LPIPS$\downarrow$ & DISTS$\downarrow$ \\
\midrule
LVSM & \cellcolor{third}29.15 & \cellcolor{third}0.9589 & \cellcolor{third}0.0395 & \cellcolor{third}0.1282 \\
LaCT-NVS & 23.65 & 0.9253 & 0.1305 & 0.2957 \\
Token-Mem & \cellcolor{second}30.83 & \cellcolor{second}0.9673 & \cellcolor{second}0.0317 & \cellcolor{second}0.1109 \\
Ours & \cellcolor{best}34.18 & \cellcolor{best}0.9786 & \cellcolor{best}0.0196 & \cellcolor{best}0.0739 \\
\bottomrule
\end{tabular}
}
\vspace{-5pt}
\caption{\textbf{Category-Agnostic Object-Centric NVS Test}. (Evaluated on scenes of dynamic objects; metrics averaged over all 30s of 30 FPS frames). Because LaCT-NVS and Token-Mem update every frame, they are prone to memory drift when evaluated at 30FPS, with LaCT-NVS degrading severely. In contrast, NSTM ranks best across all metrics, leading Token-Mem by 3.35 dB, LVSM by 5.03 dB, and LaCT-NVS by 10.53 dB in PSNR. The \colorbox{best}{best}, \colorbox{second}{second-best}, and \colorbox{third}{third-best} results are highlighted.}
\label{tab:object_generalization_results}
\vspace{-15pt}
\end{table}

\begin{figure*}[t]
  \vspace{-5pt}
  \centering
  \includegraphics[width=\textwidth]{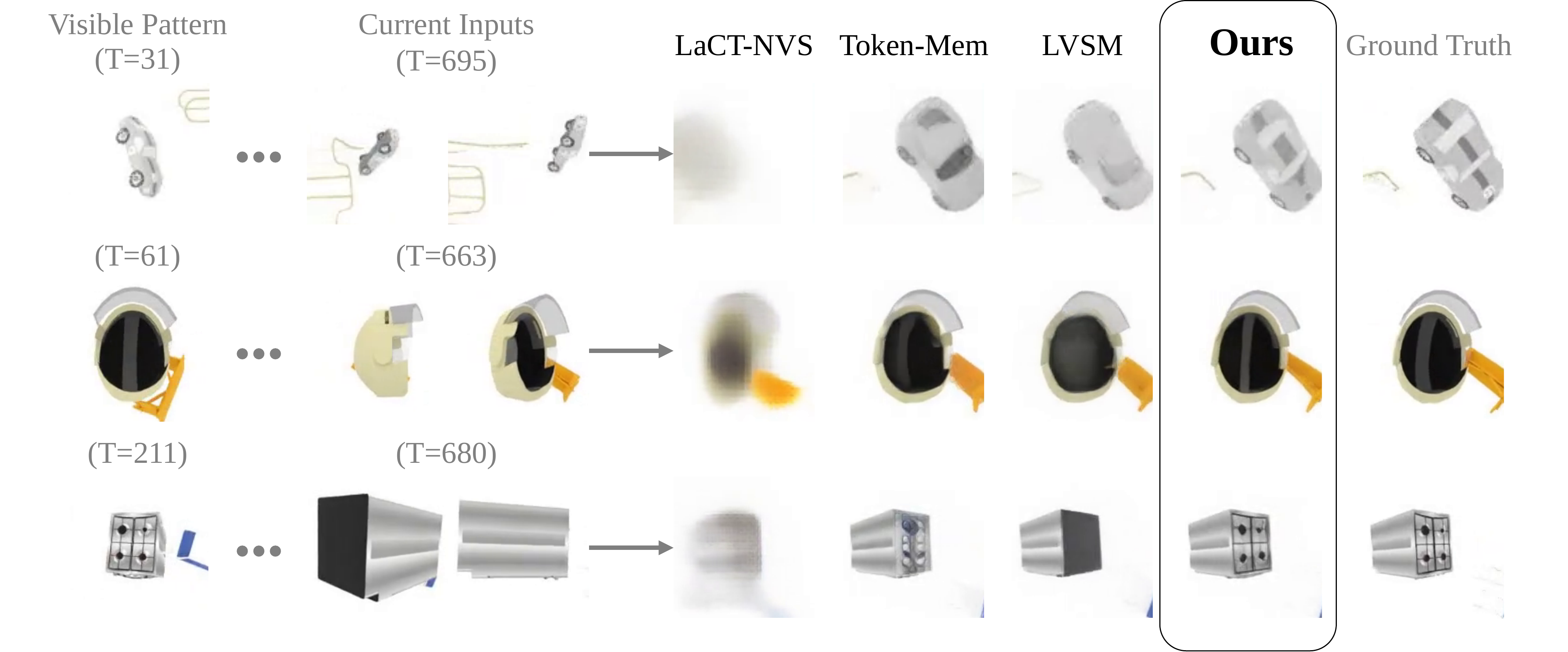} 
  \caption{\textbf{Category-Agnostic Object-Centric NVS Qualitative Results}. 
  We demonstrate our method's prior-free nature by training and evaluating on a category-agnostic dynamic object-centric dataset. \textbf{Top:} The race car stripe is memorized at $T=31$ and recovered by NSTM at $T=695$ after 22 updates, while LVSM synthesizes generic color. Token-Mem's memory degrades and LaCT-NVS drifts out of distribution. Note that NSTM is successful at recovering this pattern even with the input poses far-away from the camera while the synthesized target view is close-up. \textbf{Middle:} NSTM recovers the occluded stripe inside the helmet on frame $T=663$ that was memorized at $T=61$, while LaCT-NVS again drifts out of distribution and Token-Mem produces blurry results due to capacity constraints.  \textbf{Bottom:} Only NSTM can recall the textured pattern at $T=680$ (memorized at $T=211$). }
  \label{fig:object_generalization}
  \vspace{-5pt}
\end{figure*}

\vspace{-15px}
\noindent\textbf{Evaluation Protocol.} At every frame, we use the scene's two fixed input views and synthesize its fixed wide-baseline target view of the dynamic scene. Occlusions arise from two sources. First, the secondary object repeatedly passes in front of and behind the primary object, so regions visible from the target camera are frequently hidden from the input cameras. Second, because the primary object rotates continuously, the surface facing the target camera at any moment was seen by the input cameras earlier in the rotation and must be recalled rather than observed. Unlike our MVHumanNet++ protocol, which is limited to 1.2 FPS by the dataset's annotations (\cref{sec:inference_protocol}), all methods here run on native 30 FPS video and synthesize every frame. NSTM memorizes every 30 frames (once per second), i.e., 30 memory updates over 900 frames, mirroring the deployment schedule of \mainref{fig:teaser} (1 FPS memorization, 30 FPS synthesis). Because the objects are in constant motion, NSTM's memory is up to 29 frames stale at synthesis time, so the test also stresses its ability to reconcile a lagging memory with the current motion (see \mainref{subsec:decoupled_mem}). LaCT-NVS and Token-Mem update their memory at every frame, their native update frequency, i.e., 900 updates over the same sequence, and therefore never operate on stale memory. Metrics are averaged over all 900 frames.

\noindent\textbf{Results and Discussions.}
\cref{tab:object_generalization_results} reports quantitative results and \cref{fig:object_generalization} shows qualitative examples. NSTM ranks best on every metric, leading LVSM by 5.03 dB, Token-Mem by 3.35 dB, and LaCT-NVS by 10.53 dB in PSNR. Beyond confirming that NSTM is not human-specific, this test exposes a failure mode that our MVHumanNet++ evaluation cannot: its annotations limit us to 60 frames at 1.2 FPS, so a per-frame memory performs only 60 updates per sequence, whereas on native 30 FPS video LaCT-NVS performs 900 updates over 30 seconds (\cref{tbl:protocol_schedules}) and collapses under this regime even at its native per-frame update rate (which was its best setting in \cref{sec:sup_update_interval_analysis}). Token-Mem trails NSTM, remaining capacity constrained. NSTM's performance demonstrates the benefits of our periodic memorization design and shows that our cross-view attention successfully reconciles stale memory with current motion.

\section{Additional Implementation Details}
\label{sec:supp_implementation}

In this section, we provide comprehensive details regarding our training curriculum, optimization hyperparameters, and baseline configurations.

\subsection{Training Curriculum}
\label{subsec:sup:two_stage_curriculum}
All $256\times 256$ stateful models are trained with a two-stage training curriculum. Pretraining bootstraps our model's memory and is carried out on short video clips consisting of 4 frames.  We then finetune for \emph{free-viewpoint synthesis} and \emph{long memory persistence} with long 24-frame video clips. To demonstrate our memory retention capabilities at $512\times 512$, we further finetune our $256 \times 256$ model in stage 3 on $512 \times 512$ videos.
At each timestep, the network processes 2 sparse input views to synthesize 1 target view. Details on each stage follow:

\noindent\textbf{Stage 1: Memory Bootstrap Pretraining.} The objective of this initial phase is to explicitly encourage the network to build and rely upon its memory module. For the first frame of the sequence ($t=0$), the input consists of one frontal view (\texttt{cam00}) and one back-facing view (\texttt{cam08}), providing full geometric context. For subsequent frames ($t>0$), the inputs transition to two random views. Crucially, the target supervision is a target view that is explicitly distinct from the $t>0$ inputs. Note these target viewpoints remain the same across the sequence. This rigid setup encourages memorizing the initial context to faithfully inpaint the targets at later timesteps.

\noindent\textbf{Stage 2: Free-Viewpoint and Long Memory Adaptation Finetuning.} Once the memory module is bootstrapped, we adapt the network to encourage longer memory persistence and enable synthesis of arbitrary, time-varying viewpoints. We extend the pretraining sequence from 4 frames to 24 frames and supervise the network at each timestep by a random, time-varying target view strictly withheld from the input stream. To construct the inputs, two random base cameras are selected and held constant for all $t > 0$. At the initial timestep ($t = 0$), we enforce a forced-occlusion scenario by exclusively sourcing inputs from two distinct, alternative cameras. Finally, to ensure the model properly learns object-centric rotation rather than just camera translation, 50\% of the training sequences undergo coordinate alignment. Specifically, we modify the extrinsics of the first initial $t = 0$ camera to match the first base view. Throughout this extended sequence, we maintain the alternating memorization and synthesis steps utilized in the pre-training phase.

\noindent\textbf{Stage 3: High-Resolution ($512\times512$) Finetuning.} We further finetune at $512\times512$ resolution for 20K iterations. Similar to Stage 2, we simulate memory assimilation by injecting two alternative cameras. However, we inject the two alternative cameras at a random memorization step (even steps) rather than strictly at $t=0$. To ensure sufficient subsequent frames remain for supervision, $t$ is sampled from a decaying distribution:
$$p(t) = \begin{cases} 
  \frac{\gamma^{t/2}}{\sum_{j=0}^{T/2-1} \gamma^j} & \text{if } t \bmod 2 = 0 \\[1em] 
  0 & \text{otherwise} 
\end{cases}$$
where $\gamma=0.5$, and $T=24$ denotes the video sequence length. The $512 \times 512$ model is trained on 128 B200 GPUs.

\subsection{Training Hyperparameters}
\label{subsec:sup:main_training_hyperparams}
We implement our framework using JAX. Models are trained with a total batch size of $128$ for $500,000$ iterations during the pretraining stage, followed by an additional $25,000$ iterations during the finetuning stage. The $512 \times 512$ model undergoes another stage of high-resolution finetuning as in \cref{subsec:sup:two_stage_curriculum}. We employ a cosine learning rate decay schedule with a linear warmup. During pretraining, the learning rate is initialized at $1 \times 10^{-6}$, warms up over $4,000$ steps to a peak of $4 \times 10^{-4}$, remains constant for $100,000$ steps, and then decays down to a final value of $1 \times 10^{-6}$. During the finetuning stage, we retain the same schedule parameters but reduce the peak learning rate to $4 \times 10^{-5}$ warming up over $4,000$ steps and then remaining constant. Finally, for the image reconstruction objective across all models and stages, the perceptual loss weight is fixed at $\lambda_{\text{lpips}} = 0.5$.

\subsection{Training for the Object-Centric NVS Test}
\label{sec:supp_object_training}

\noindent\textbf{Training Curriculum.} For the object-centric experiment
(\cref{subsec:sup:sec:object_centric_nvs}) we follow the two-stage curriculum of \cref{subsec:sup:two_stage_curriculum}, simplified by the dataset's
camera setup: every scene provides exactly two input cameras and one
target camera, randomly sampled once per scene and fixed
over the sequence. We inject no  forced-occlusion views at $t=0$, and the target view is held fixed, so Stage~2 (24-frame clips) serves purely as long-memory
adaptation. Training clips are sampled from the 6 FPS training
sequences with random strides. We still train on 4-frame sequences for stage 1 and 24-frame sequences for stage 2.

\noindent\textbf{Training Hyperparameters.} All models are pretrained with a batch size of 128 for 120{,}000 iterations with a peak learning rate of $4 \times 10^{-4}$. %
The stateful models (NSTM, LaCT-NVS, and Token-Mem) are then finetuned for 16{,}000 iterations in Stage~2 with a peak learning rate of $1 \times 10^{-4}$; as in \cref{sec:sup:baseline}, the stateless LVSM is trained only on the final frame of the 4-frame stage-1 clips.

\begin{table*}[t]
\centering
\resizebox{\textwidth}{!}{%
\begin{tabular}{llccccccc}
\toprule
& & & & \multicolumn{2}{c}{NSTM} & \multicolumn{2}{c}{LaCT-NVS / Token-Mem} & \\
\cmidrule(lr){5-6} \cmidrule(lr){7-8}
Dataset & Protocol & Frame rate & Timesteps & Update interval & \# updates & Update interval & \# updates & Synth.\,:\,update \\
\midrule
\multirow{5}{*}{MVHumanNet++}
 & Training, Stage 1            & 1.2 FPS & $T{=}4$   & 2 frames & 2  & 1 frame & 4  & 1:1 \\
 & Training, Stage 2 / 3        & 1.2 FPS & $T{=}24$  & 2 frames & 12 & 1 frame & 24 & 1:1 \\
 & Evaluation (Stress, Natural, NVS, Ablation) & 1.2 FPS & $T{=}60$ & 2 frames & 30 & 1 frame & 60 & 1:1 \\
 & Update-Interval Analysis (\cref{sec:sup_update_interval_analysis}), $\tau \in \{1,2,4,8,16,32\}$ & 1.2 FPS & $T{=}60$ & $\tau$ frames & $60/\tau$ & $\tau$ frames & $60/\tau$ & $(\tau{-}1)$:1 \\
 & Inference / rendering (supp.\ 30 FPS videos) & 30 FPS & $T{=}1800$ (60 s) & 30 frames (1 s) & 60 & 1 frame (33 ms) & 1800 & 29:1 \\
\midrule
\multirow{3}{*}{\begin{tabular}[c]{@{}l@{}}Category-Agnostic\\Objects \end{tabular}}
 & Training, Stage 1            & $6/s$ FPS, random $s$ & $T{=}4$   & 2 frames & 2  & 1 frame & 4  & 1:1 \\
 & Training, Stage 2            & $6/s$ FPS, random $s$ & $T{=}24$  & 2 frames & 12 & 1 frame & 24 & 1:1 \\
 & Evaluation (\cref{subsec:sup:sec:object_centric_nvs}) & 30 FPS & $T{=}900$ (30 s) & 30 frames (1 s) & 30 & 1 frame (33 ms) & 900 & 29:1 \\
\bottomrule
\end{tabular}}
\caption{\textbf{Memorization and synthesis schedules across all protocols.} We summarize the schedules of all protocols described in our paper. Constrained by the annotation provided by MVHumanNet++~\cite{li2025mvhumannet++}, we train and evaluate with 1.2 FPS frames. This protocol fairly demonstrates our long-horizon quality, as our synthesis steps operate independently in between memorization steps. However, it provides an advantage to baselines without a decoupled strategy, as the lower frame rate means their memory will not degrade as much as it would at 30 FPS. Therefore, we further show video comparison results on real minute-long 30 FPS videos 
on our supplementary website. 
For the Category-Agnostic Object-Centric NVS Test, training clips are sampled from the 6 FPS training sequences with a temporal stride $s$ drawn at random per clip (\cref{sec:supp_object_training}), while the evaluation is done on 30 FPS videos.}
\label{tbl:protocol_schedules}
\end{table*}

\subsection{Baseline Adaptations and Coordinate Alignments}
\label{sec:sup:baseline}
To ensure a fair comparison, all baselines evaluated 
are of $256 \times 256$ resolution and all stateful models are trained on the exact same two-stage training curriculum.
The LVSM baseline~\cite{jin2025lvsm} is stateless and thus goes through only one pretraining stage as below.
Due to architectural differences, camera coordinate systems must be aligned differently:

\noindent\textbf{LVSM~\cite{jin2025lvsm}:} We utilize the exact architecture described in the original LVSM paper. But for fair comparison, we use the same learning rate schedule as above and the same L2 and LPIPS loss. Because LVSM is a stateless, per-frame model, we construct its training batches by extracting only the final frame ($t=3$) from our 4-timestep pretraining video clips, ensuring it trains on identical data distributions. For coordinate alignment, we relativize the camera coordinate system to each individual target camera over time (i.e., the specific target camera for that frame is set as the origin with an identity rotation).

\noindent\textbf{LaCT-NVS~\cite{zhang2025test_lact} and NSTM (Ours):} Stateful models must maintain coherent camera coordinate system across time and adapt to time-varying target cameras. Therefore, for LaCT and NSTM, we relativize all cameras in the sequence to the \textit{first} target camera ($t=0$), ensuring a stable, global coordinate space for the fast-weights to operate within. The LaCT-NVS baseline in
all comparison tables retains the original dot-product inner loss of LaCT, adapted to causal per-frame updates for streaming; a variant equipped with our $L_2$ inner loss is analyzed in the ablation study in \mainref{tab:ablations}.

\noindent\textbf{Token-Mem~\cite{cut3r}:} As explained in \mainref{subsec:setup}, \emph{Token-Mem} is a custom novel view synthesis model whose memory mechanism is based on CUT3R~\cite{cut3r}. As a stateful method, it also requires a temporally consistent coordinate frame, and we use the same coordinate system as described above for LaCT and our NSTM. Architecture-wise, we follow the original CUT3R~\cite{cut3r} implementation to use a state size of 768 tokens each with 768 feature channels. Unlike the pose-free CUT3R, Token-Mem operates on posed inputs: following LVSM and NSTM, input images are channel-wise concatenated with their per-pixel Pl\"ucker ray maps before patchification, and target views are specified by patchified Pl\"ucker rays of the target camera. Also differing from CUT3R, we optimize L2 and LPIPS losses consistent with the other methods compared.

\subsection{Evaluation Protocol Details}
\label{sec:supp_eval_protocol_details}
\textbf{Evaluation Dataset Filtering.} Across all our long-term evaluation protocols---including 
the Memory Stress Test (\mainref{subsec:results}), the Ablation Study (\mainref{sec:Ablation}),
the Memory from Natural Rotations Test (\cref{sec:supp_natural_memory_test}), and the General Novel View Synthesis Test (\cref{sec:nvs})---we stress-test model performance over extended time horizons (i.e., up to $t=60$). Consequently, we filtered the originally provided MVHumanNet++ test set to strictly include sequences containing at least 60 frames. This curation resulted in a unified subset of 390 evaluation scenes, which differs from the standard MVHumanNet++ test split but is applied consistently across all our long-term evaluations.

\noindent\textbf{Sequence Ordering for Natural Rotations}. For the Natural Rotations Test
(\cref{sec:supp_natural_memory_test}), we utilize 17 scenes featuring natural back-revealing motions. Because these rotations typically occur near the end of the raw videos, we temporally reverse the sequences. This ensures the unoccluded back is observed early, allowing us to rigorously assess long-term memory retention (up to $t=60$) over the subsequent minute-scale horizon.

    \begin{figure*}[t]
\vspace{-15pt}
\includegraphics[width=1.0\textwidth]{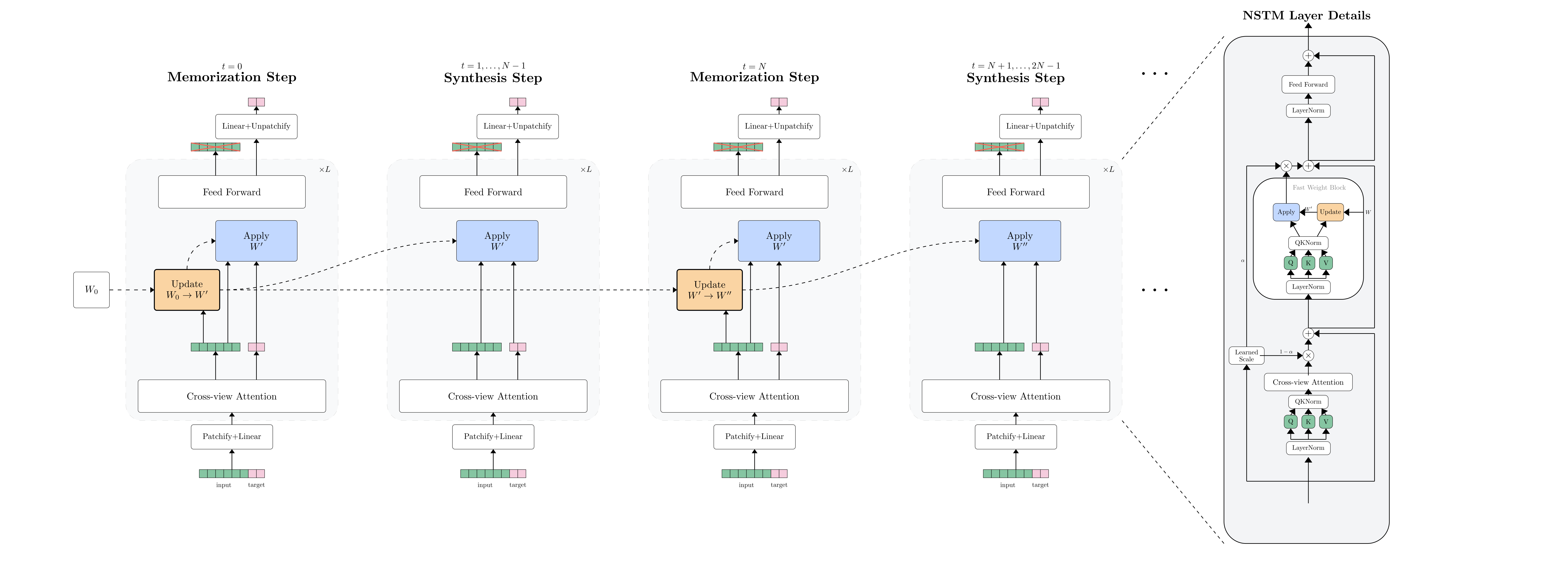}
\vspace{-20pt}
\caption{\textbf{Online Inference Pipeline and NSTM Layer Design.} \textbf{Left}: During inference, we run the memorization step periodically while performing the synthesis step per-frame, which enables us to achieve amortized real-time speed. \textbf{Right}: We introduce the NSTM Layer consisting of cross-view attention, a fast-weight block and a feed-forward layer. We also incorporate a learned scale that allows the network to balance the contribution of memory and attention to each layer's output.}
\label{fig:nstm_model_block_architecture}
\vspace{-10pt}
\end{figure*}

\subsection{Online Inference Protocol}\label{sec:inference_protocol}

\noindent\textbf{Decoupled Memorization and Synthesis Frequencies.} Unlike models that rigidly couple memory updates and queries at every timestep, NSTM decouples memorization and synthesis strides. \cref{tbl:protocol_schedules} summarizes the schedules used throughout the paper. MVHumanNet++~\cite{li2025mvhumannet++} releases its annotated train and evaluation sets only at 1.2 FPS and this constraint fixes the temporal resolution at which quantitative evaluation on MVHumanNet++ is possible. On these sequences we train and evaluate by memorizing every other frame, following our decoupling strategy. With the exception of the \emph{Update Interval Analysis}, all our quantitative MVHumanNet++ results in
\cref{sec:experiments} and \cref{sec:supp_natural_memory_test,sec:nvs}
 employ this schedule, whereas the baselines update at every frame. To isolate the effect of the update frequency (\cref{sec:sup_update_interval_analysis}), we further analyze all methods--including ours -- under an equalized memory update interval $\tau$. The schedule of 1 FPS memorization and 30 FPS synthesis 
 (\mainref{fig:teaser})
 cannot be scored on MVHumanNet++ but is qualitatively demonstrated in our supplementary website videos. These videos are rendered from the raw minute-long 30 FPS MVHumanNet++ sequences, requiring only 60 memory updates for NSTM compared to 1,800 per-frame updates for the baselines. Finally, because our Category-Agnostic Object-Centric NVS test is free from these frame rate constraints, it is quantitatively evaluated on native 30 FPS videos (\cref{subsec:sup:sec:object_centric_nvs}).

\subsection{Detailed Model Architecture}
\label{sec:detailed_architecture}

\cref{fig:nstm_model_block_architecture} shows our NSTM layer in more details.  During inference, we run the memorization step periodically while performing the synthesis step per-frame. Different from training time, there are no memory tokens and we extract the target tokens as output for all steps. More specifically, following LVSM~\cite{jin2025lvsm}, we channel-wise concatenate input RGB images with their corresponding per-pixel Pl\"ucker ray maps, using linear layers to patchify them into input tokens. Similarly, we patchify the Pl\"ucker rays of the target camera to form target tokens. These concatenated input and target tokens are then passed through a $L=24$ Transformer blocks, each comprising a cross-view attention module, a TTT memory module, and a feed-forward MLP module. The cross-view attention module applies self-attention on the concatenated input and target tokens. Beyond a standard TTT layer, we introduce a learned weight $\alpha$ to balance the contribution from the \emph{cross-view attention} module and the \emph{TTT} memory module. Finally, we extract the processed target tokens and unpatchify them with a linear layer, applying a sigmoid activation to predict the target RGB images.

\subsection{Foreground Matting and Background Augmentation}
\label{subsec:foreground_matting}

To facilitate clean isolation of the dynamic subject from the studio background, we output a 4-channel $\text{RGB}\alpha$ image. During training, we employ background augmentation: we composite both the ground-truth foreground and our model's predicted $\text{RGB}\alpha$ output onto randomly sampled solid color backgrounds before computing the photometric loss. This augmentation forces the network to learn a highly accurate alpha matte. Consequently, at inference time, NSTM produces a clean, disentangled alpha channel, enabling seamless compositing of the synthesized human subject into arbitrary novel environments.

\begin{figure*}[t]
  \centering
  \includegraphics[width=\textwidth]{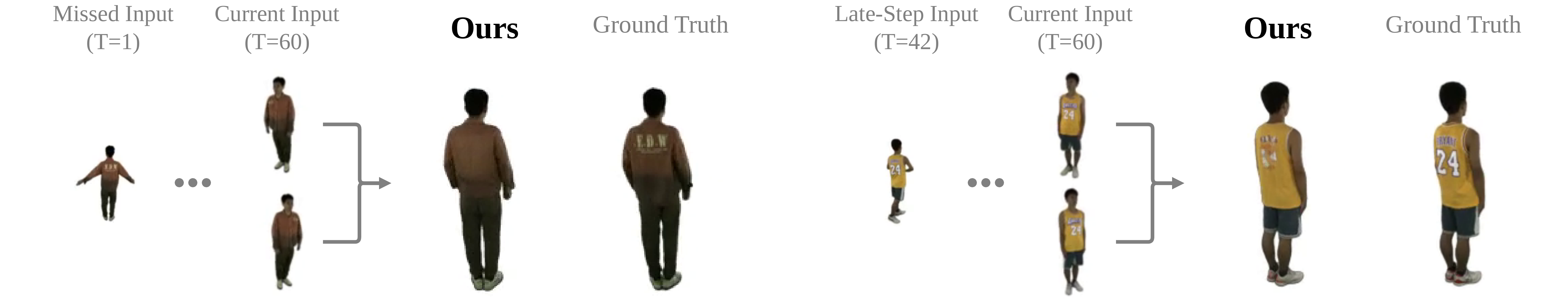}
  \caption{\textbf{Limitations}. 
  \textbf{Left: Update Frequency and Missed Events.} Our periodic updates may miss events between memorization steps. The pattern shown at $t=1$ was presented only at this non-update timestep and is therefore not recoverable at $t=60$. \textbf{Right: Memory Primacy Bias} Our method favors memorization during early steps. Here the back is revealed after 21 updates (at $t=42$) and our recovery at $t=60$ is only partial.}
  \label{fig:limitations_fig}
  \vspace{-10pt}
\end{figure*}

\section{Limitations}
\label{sec:supp_limitations}
While NSTM provides a principled solution for streaming dynamic NVS from multi-view videos, we identify several avenues for future research. \emph{Update Frequency and Missed Events:} Periodic updates may miss events between memorization steps. Large-chunk updates that memorize multiple frames at once, each with its own time embedding, could mitigate this and additionally support irregular frame spacing (e.g., unsynced multi-view rigs), leaving open the tradeoff between update frequency and chunk size. \emph{Finite Memory Capacity:} Models like ours eventually hit capacity limits; future iterations could manage memory caching weights in a learned way. \emph{Memory Primacy Bias:} The method favors memorization during early steps; larger-scale training on longer sequences and better mid-sequence frame sampling could encourage long-term retention. \emph{Complex Motion and Trajectories:} Resolving long-range camera and subject motion within the compressed memory remains challenging and is left for future work. See \cref{fig:limitations_fig} for limitation examples.

\section{Concurrent Work Discussion}
\label{sec:supp_concurrent}
In this section we discuss two concurrent works that utilize the fast-weight layers NSTM builds on; neither targets online synthesis from multi-view streams.

\noindent\textbf{TTT as linear attention.}
Liu~\emph{et~al.}~\cite{liu2026tttla} show that TTT layers with a KV-binding inner loss and a linear output layer can be rewritten as learned linear attention, and that several inner-loop components can be removed to obtain a parallel form with up to $4\times$ higher TTT-layer throughput, validated with a single chunk update on static scenes. Their analysis is orthogonal to our contributions---the decoupled memorization/synthesis schedule, memory supervision, and memory stability over 30--60 sequential updates---and we leave adopting their simplifications to reduce our memorization cost to future work.

\noindent\textbf{Fast Spatial Memory (FSM).}
Ma~\emph{et~al.}~\cite{ma2026fsm} stabilize multi-chunk LaCT for \emph{offline} 4D reconstruction with an EWC-style consolidation that pulls the fast weights toward an exponential-moving-average anchor after each chunk update. FSM buffers 32 posed frames, processes them in four chunks (four fast-weight updates per scene), and renders view--time queries offline. In contrast, NSTM targets \emph{online} streaming scenarios where latency is critical. Our cross-view attention plays a vital role here by reconciling ongoing motion with stale memory. Furthermore, our decoupling strategy, combined with an $L_2$ inner loss and memory caching, successfully maintains a stable memory over minute-long, 1,800-frame videos with 60 updates.

\end{document}